\documentclass[9pt,cspaper,compsoc]{IEEEtran-technote}

\ifCLASSOPTIONcompsoc
\else
   \usepackage{cite}
\fi

\ifCLASSINFOpdf
  \usepackage[pdftex]{graphicx}
  \graphicspath{{./pdf/}{./jpeg/}{./Figs/}{./NewFigs/}{./author_Photo/}}
\else
\fi

\usepackage[cmex10]{amsmath}

\ifCLASSOPTIONcompsoc
\usepackage[font=footnotesize]{caption}

\usepackage{times}
\usepackage{graphicx}
\usepackage{amsmath}
\usepackage{amssymb}

\usepackage{amsfonts}
\usepackage{mathrsfs,xspace}
\usepackage{cite}

\usepackage{mathrsfs}
\usepackage[lined,boxed]{algorithm2e}

\usepackage{setspace}

\newcommand{\eighthao}{\fontsize{9pt}{9pt}\selectfont}

\def\mathbbold{\mathbf}

\def\KNN{$k$NN\xspace}

\renewcommand{\vspace}[1]{{}}

\begin{document}

\title{Context-Aware Hypergraph Construction for Robust Spectral Clustering}

\author{ \eighthao Xi Li, Weiming Hu, Chunhua Shen, Anthony Dick, Zhongfei Zhang  %
\IEEEcompsocitemizethanks
{
\IEEEcompsocthanksitem X. Li and W. Hu are with National Laboratory of Pattern Recognition, Institute of Automation, Chinese Academy
of Sciences, China.
E-mail: $\{lixi, wmhu\}$@nlpr.ia.ac.cn
\IEEEcompsocthanksitem X. Li, C. Shen, and A. Dick are with School
of Computer Science, The University of Adelaide, Australia.
\IEEEcompsocthanksitem Z. Zhang is with State University of New York, Binghamton, NY 13902, USA.
}
}

\markboth{Appearing in IEEE Transactions on Knowledge and Data Engineering, 2013}{ }

\IEEEcompsoctitleabstractindextext{%
\begin{abstract}

    Spectral clustering is a powerful tool for unsupervised data analysis. In this paper,
we propose a context-aware hypergraph similarity measure (CAHSM), which leads to robust
spectral clustering in the case of noisy data. We construct three types of
    hypergraph---the pairwise hypergraph, the $k$-nearest-neighbor
    (\KNN) hypergraph, and the high-order over-clustering hypergraph.
    The pairwise hypergraph captures the pairwise
    similarity of data points; the \KNN hypergraph captures the
    neighborhood of each point; and the clustering
    hypergraph encodes high-order contexts within the dataset.  By
    combining the affinity information from these three
    hypergraphs, the CAHSM algorithm is able to
    explore the intrinsic topological information of the dataset.
    Therefore, data clustering using CAHSM tends to be more robust.
    Considering the intra-cluster compactness and the inter-cluster
    separability of vertices, we further design a discriminative
    hypergraph partitioning criterion (DHPC).
    Using both CAHSM and DHPC, a robust spectral clustering algorithm
    is developed.  Theoretical analysis and
    experimental evaluation demonstrate the effectiveness and
    robustness of the proposed  algorithm.

\end{abstract}


}

\maketitle

\IEEEdisplaynotcompsoctitleabstractindextext

\IEEEpeerreviewmaketitle

\begin{spacing}{1.00}

\section{Introduction}
\label{sec:intro}

Spectral clustering is an effective means of clustering data with complex topological structure~\cite{Ng-Jordan-Weiss15,Zelnik-Manor-Perona-NIPS2005,Li-Liu-Chen-Tang-ICCV2007,Gdalyahu-Weinshall-Werman7,
Yu-Shi14,Von-Luxburg-SC2007,Nadler-Lafon-Coifman-Kevrekidis-NIPS2006,Ding-He9,lu2008constrained,yang2012discriminative,wang2009clustering,cai2007spectral}.
It plays an important role in unsupervised learning from data,
and therefore has a wide range of applications, including
circuit layout~\cite{Alpert-Kahng1,Chan-Schlag-Zien2}, load
    balancing~\cite{Hendrickson-Leland3}, image
    segmentation~\cite{Shi-Malik4,Malik-Belongie-Leung-Shi5,
    Weiss-ICCV1999, Meila-Shi-NIPS2000,Chang-Yeung-ICCV2005},  motion
    segmentation~\cite{Lauer-Schnorr-iccv2009}, video
    retrieval~\cite{Hu-Xie-TIP2007}, etc.
Typically, the affinity relationships between data samples are modeled by a graph,
and therefore spectral clustering aims to optimize a graph partitioning criterion
for data clustering
based on local vertex similarities.
    However, there are still several unsolved issues for
    traditional spectral clustering methods:
    i) how to automatically discover the number of clusters;
    ii) how to correctly choose the scaling parameter for graph construction;
    iii) how to counteract the adverse effect of noise or outliers; and
    iv) how to incorporate different types of information to enhance the
    clustering performance.

    In the literature, Zelnik-Manor and Perona
    \cite{Zelnik-Manor-Perona-NIPS2005} attempt to address issues
    i) and ii) by designing a local scaling mechanism, which
    adaptively calculates the affinity matrix and explores the
    intrinsic structural information on the energy eigenvalue spectrum
    of the normalized graph Laplacian to discover the number of
    clusters.  However, this local scaling mechanism is
    susceptible to noise or outliers.  Following~\cite{Zelnik-Manor-Perona-NIPS2005}, Li \emph{et
    al.}~\cite{Li-Liu-Chen-Tang-ICCV2007} propose a noise robust spectral
    clustering (NRSC) algorithm to resolve issues i) and iii).
    The proposed NRSC algorithm can automatically estimate the number
    of clusters via computing the largest eigenvalue gap of the
    normalized graph Laplacian.  In addition, the proposed NRSC
    algorithm maps the
    original data samples (vertices) into a new space, in which the
    clusters have a higher intra-cluster compactness and inter-cluster
    separability. If the noisy data samples are weakly
    interconnected with each other, this mapping relocates
    the noisy data samples around the origin of the new
    space.  This usually results in a compact noise cluster.  However,
    the real-world data samples (e.g., images and videos) within a
    cluster are often not densely interconnected due to problems with the visual feature description.  Therefore,
    the mapping may lead to topological information loss
    for the clusters.  As a result, the weakly interconnected samples
    in the ordinary clusters are also relocated around the origin of the new space. This may result in  low separability between clusters.

    More recently,  hypergraph
    analysis~\cite{Zhou-Huang-Schokopf-nips2006,Agarwal-Lim-Manor-Perona-Kriegman-Belongie-CVPR2005}
    has emerged as  a popular tool for addressing issues iii) and iv).
    The fundamental idea of hypergraph analysis is
    to explore the underlying affinity
    relationships among vertices by constructing a hypergraph with a variety of
    hyperedges that capture affinity. This has been applied to many domains
    such as image matching~\cite{Zass-Shashua-CVPR2008}, multi-label
    classification~\cite{Sun-Ji-Ye-KDD2008}, video object
    segmentation~\cite{Huang-Liu-Metaxas-CVPR2009}, and image
    retrieval~\cite{Huang-Liu-Zhang-Metaxas-CVPR2010}.
    For example, Sun \emph{et al.}~\cite{Sun-Ji-Ye-KDD2008} carry out hypergraph construction
    by sequentially introducing new vertices into existing hyperedges using
    clique expansion or star expansion.
    Huang \emph{et al.}~\cite{Huang-Liu-Zhang-Metaxas-CVPR2010} propose a probabilistic hypergraph
    model that softly assigns a vertex to a hyperedge according to the similarity between
    the vertex and the centroid of the hyperedge.

    \textbf{Motivation and contribution}
    In general, most existing spectral clustering algorithms only
    focus on the pairwise interactions between vertices.  In other
    words,
    the pairwise similarity between two vertices is only based on the
    individual vertices themselves.  If a vertex is corrupted,
    this pairwise similarity can change significantly. Consequently,
    their true affinity may not be stably represented.  Thus,
    designing a robust similarity measure is one of the key problems
    in data clustering.

    Here,  we show that the high-order contextual information on
    vertices can help alleviate this problem.
Contexts are groups of vertices that share some common properties.
Once contexts have been computed, the vertex similarity
    measure depends on not only two individual vertices but also their
    corresponding contexts.  The similarity measure that includes contextual
    information is much more stable because it takes into account local
    grouping and neighborhood information of each vertex.  When a single
    vertex is corrupted, the high-order contextual
    similarity can still provide complementary information to
    counteract the impact of the corruption.

    Motivated by this
    observation, we
    propose a robust spectral clustering algorithm based on a
    context-aware hypergraph similarity measure. We use three
    different types of hypergraphs: pairwise hypergraph,
    $k$-nearest-neighbor
    (\KNN) hypergraph, and high-order over-clustering hypergraph.
    The pairwise hypergraph is capable of encoding
     pairwise affinity information on vertices.
    In contrast, the $k$NN hypergraph
    and the over-clustering hypergraph capture the
    underlying manifold structure on vertices by modeling
    their high-order neighborhood and contextual grouping properties, respectively.
    By combining these hypergraphs, we obtain the context-aware hypergraph similarity
    measure that characterizes the intrinsic connectivity relationships
    among vertices, resulting in the clustering robustness in the case of noise or outlier
    corruption.

The main contributions of this work are therefore three-fold.
\begin{itemize}
\item We introduce a high-order context into the spectral clustering process.
The high-order context of a vertex is defined as a
set of vertices with similar properties to the  vertex.
Each  vertex in a context is influenced by
other vertices in the same context.

\item The problem of building the high-order context is converted to that of hypergraph construction,
which encodes the local affinity information using different types of hypergraphs.
To this end, we design three types of hypergraphs
to encode the pairwise, neighboring, and local grouping
information on vertices.
Based on these hypergraphs, we further propose
a context-aware hypergraph similarity measure (CAHSM) to capture the intrinsic topological information on
vertices.
In essence, CAHSM is a generalization of traditional similarity measures, and aims to utilize
the hypergraph context to explore the underlying affinity relationships between vertices.

\item

    We propose a discriminative hypergraph partitioning criterion (DHPC) to
    characterize the intra-cluster compactness and the
    inter-cluster separability of vertices.  By maximizing the DHPC, we
    effectively capture the discriminative information on vertices.
    The optimization of DHPC can be relaxed into a trace-ratio
    maximization problem.
     Using both CAHSM and DHPC, we develop
    a pairwise+$k$NN+over-clustering hypergraph spectral clustering algorithm
    (referred to as PKO+HSC)
    for data clustering.

\end{itemize}

\begin{figure}[t]
\vspace{-0.1cm}
\begin{center}
\includegraphics[width=0.92\linewidth]{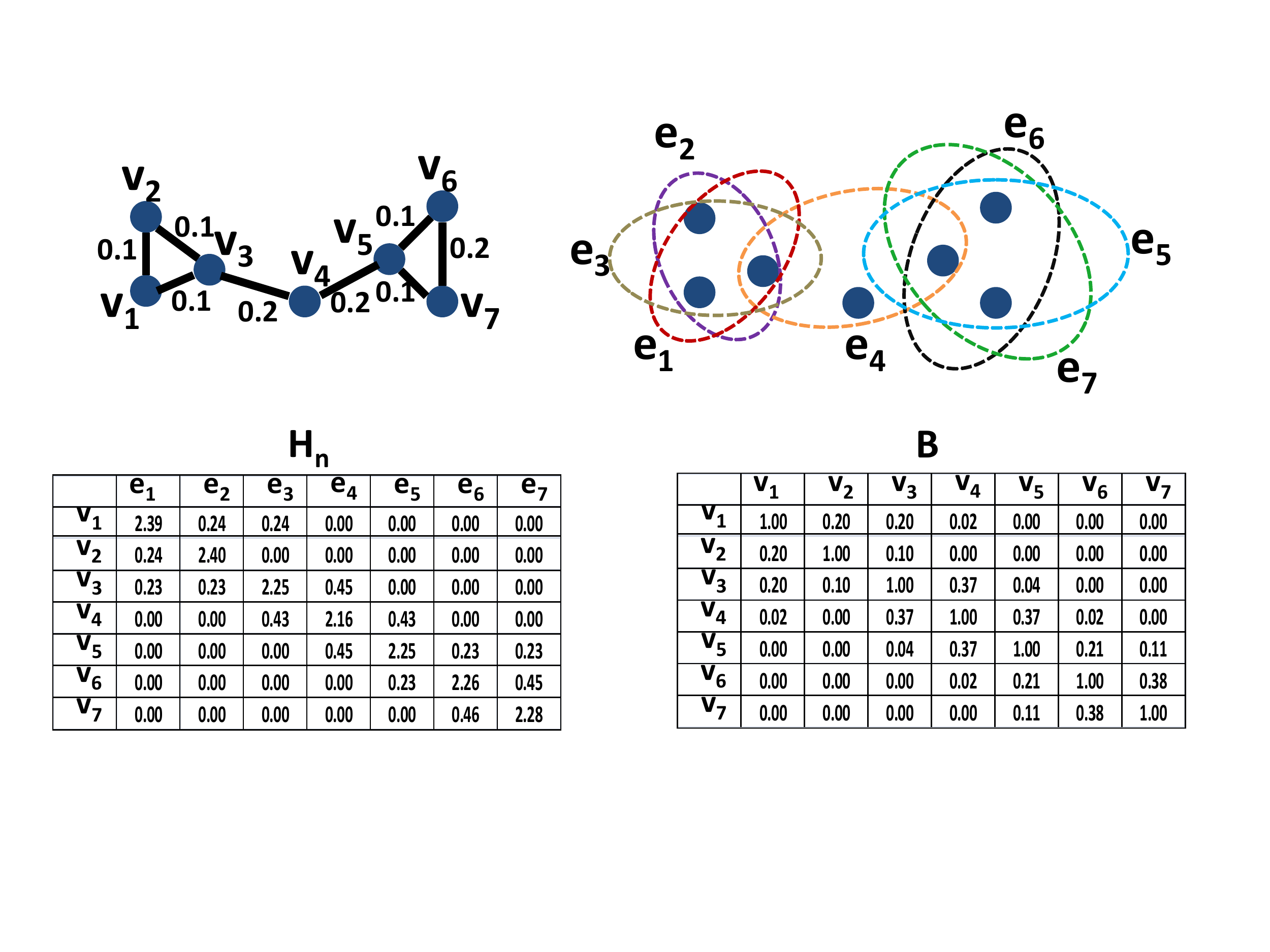}
\end{center}
\vspace{-0.35cm}
\caption{Illustration of the \KNN hypergraph construction. The top-left subfigure shows the pairwise edge
between any vertex and its two nearest neighbors; the top-right
subfigure displays the associated hypergraph structure whose
hyperedges (highlighted by ellipses of different colors) consist of
each vertex and its two nearest neighbors; the bottom-left subfigure
shows the \KNN hypergraph incidence matrix $H_{n}$; and the
bottom-right subfigure exhibits the \KNN hypergraph similarity matrix
$B$. } \vspace{-0.38cm}
\label{fig:hypergraph_demo}
\end{figure}

\section{Context-aware  hypergraph construction \label{sec:CADHL}}

In what follows, we first
discuss how to construct a robust hypergraph similarity measure
using three different types of hypergraphs, and then
describe a discriminative  hypergraph partitioning criterion for
data clustering.

\vspace{-0.2cm}
\subsection{Context-aware hypergraph similarity measure\label{sec:Prob_hypergraph}}

In order to effectively explore the high-order affinity relationships among
vertices, we propose a hypergraph construction mechanism based on
three types of hypergraphs, which are the pairwise hypergraph, the
$k$-nearest-neighbor (\KNN)
hypergraph, and the over-clustering hypergraph.  The pairwise hypergraph
reflects the pairwise relationships between vertices;
the \KNN hypergraph characterizes the neighboring information
on vertices, and the over-clustering hypergraph captures the local grouping
relationships among vertices. By combining these three types of hypergraphs,
the proposed hypergraph construction mechanism is capable of exploring the underlying high-order
affinity relationships among vertices.

\emph{1) Pairwise hypergraph.}
For easy exposition, let $\mathbb{Z}=\{\mathbf{z}_{i}\}_{i=1}^{N}$ denote a sample set.
Based on $\mathbb{Z}=\{\mathbf{z}_{i}\}_{i=1}^{N}$, we create a pairwise
graph $G_{p}$ with $N$ vertices.
Mathematically, the graph $G_{p}$ can be denoted as $G_{p}=(\mathbb{V},E_{p}, W_{p})$,
where $\mathbb{V}=\{v_{i}\}_{i=1}^{N}$ is the vertex set corresponding
to $\{\mathbf{z}_{i}\}_{i=1}^{N}$,
$E_{p}\subseteq \mathbb{V}\times \mathbb{V}$ is the edge set
containing all possible pairwise edges, and $W_{p}$ is the edge-weight
function returning the affinity
value between two vertices.
In practice, the graph $G_{p}$ is formulated as a weighted similarity
matrix
$A=(a_{ij})_{N\times N}$: \vspace{-0.15cm}
\begin{equation}
a_{ij}=\left\{
\begin{array}{cc}
W_{p}(v_{i},v_{j}) & \mbox{if}\thickspace (v_{i},v_{j})\in E_{p},\\
0 & \mbox{otherwise},
\end{array}\right. \vspace{-0.15cm}
\label{eq:pairwise_weight}
\end{equation}
where $W_{p}(v_{i},v_{j})=\mathcal{G}(\mathbf{z}_{i}, \mathbf{z}_{j})$ is a kernel function used for
measuring the similarity between $\mathbf{z}_{i}$ and $\mathbf{z}_{j}$.
Note that the above procedure of graph creation is independent of the choice of kernel functions.
In other words, it is easy to incorporate various kernel functions into the above graph creation process.

\begin{figure}[t]
\vspace{-0.2cm}
\begin{center}
\includegraphics[width=0.9\linewidth]{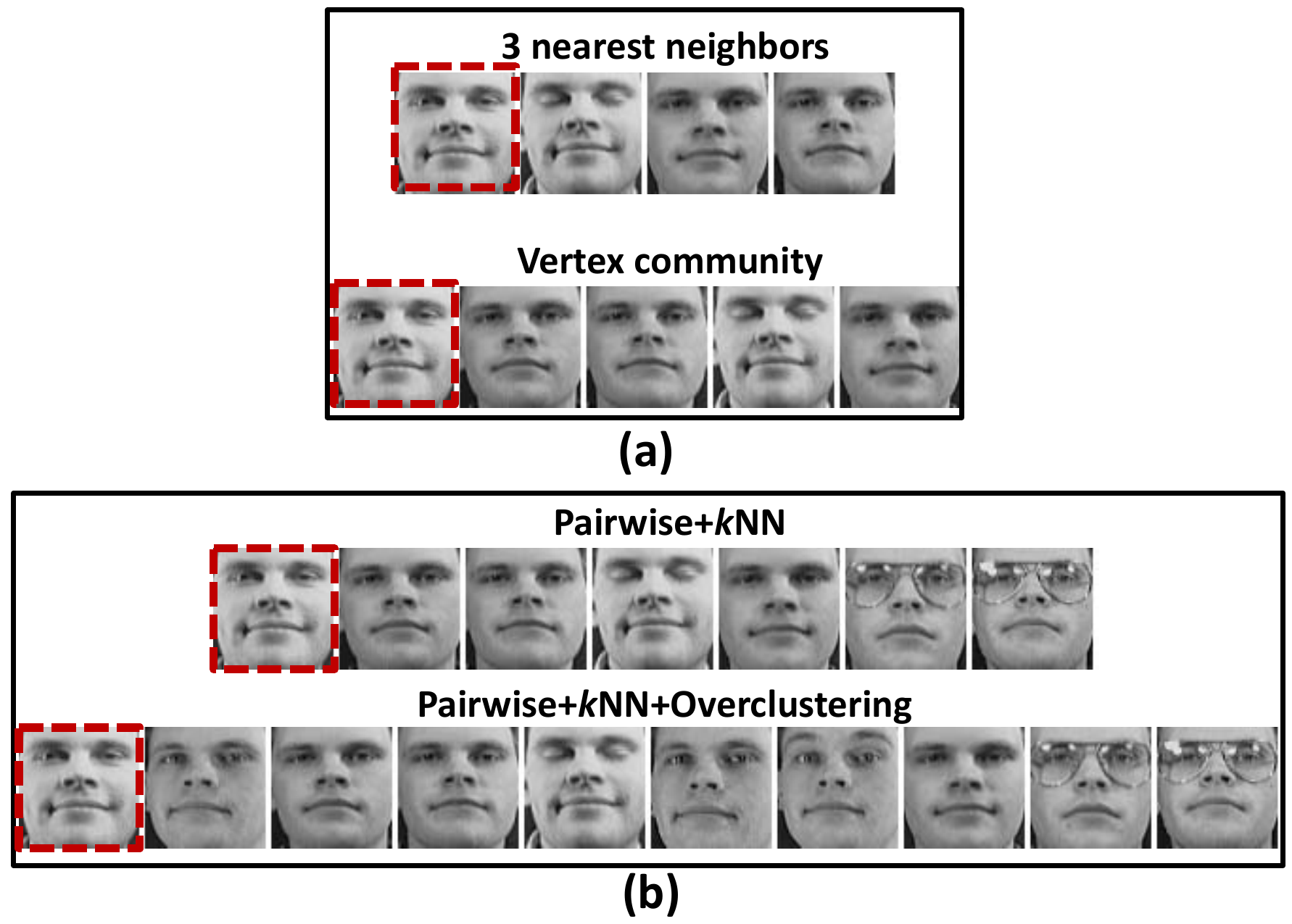}
\end{center} \vspace{-0.36cm}
\caption{Illustration of different hypergraph concepts and hypergraph clustering results based on
different hypergraph similarity measures. Specifically, the upper part of (a) shows the 3 nearest neighbors of the face image highlighted by the dot-dashed
bounding box; the lower part of (a) displays a vertex community that is a high-order context of the face image highlighted by the dot-dashed
bounding box; the upper and lower parts of (b) respectively exhibit the clustering results containing the face image (highlighted by the dot-dashed
bounding box)
using
pairwise+\KNN and pairwise+$k$NN+over-clustering hypergraph similarities (respectively corresponding to PK+HSC and the proposed PKO+HSC).
Clearly, the proposed pairwise+$k$NN+over-clustering similarity measures performs best.}
\label{fig:clustering_example} \vspace{-0.38cm}
\end{figure}

According to the hypergraph theory, the pairwise
graph $G_{p}$ is merely a special hypergraph whose hyperedge
cardinality  equals 2. Therefore, we reformulate
$G_{p}$ using the hypergraph terminologies. As a generalization of a traditional pairwise graph, a hypergraph
is composed of many hyperedges, and each hyperedge corresponds
to a set of  vertices which have some common properties.
Mathematically, these hyperedges are generally associated with a hypergraph
incidence matrix $H_{p}=(h_{p}(v_{i}, e^{p}_{\ell}))_{|\mathbb{V}|\times |E_{p}|}$: \vspace{-0.1cm}
\begin{equation}
h_{p}(v_{i}, e^{p}_{\ell}) =
\left\{
\begin{array}{ll}
1, & \mbox{if} \thickspace v_{i}\in e^{p}_{\ell},\\
0, & \mbox{otherwise},
\end{array}
\right. \vspace{-0.1cm}
\end{equation}
where $e^{p}_{\ell}=(v_{m}, v_{n})$ is the $\ell$-th hyperedge of $E_{p}$.
In order to measure the degree of the within-hyperedge
vertices belonging to the same cluster, we introduce the notion
of the pairwise hyperedge weight, which is defined as the pairwise similarity
of the vertices in each hyperedge. So
we can define the pairwise hypergraph similarity as: \vspace{-0.1cm}
\begin{equation}
u_{ij}=
\underset{e^{p}_{\ell}\in E_{p}}{\sum}\eta_{p}(e^{p}_{\ell})h_{p}(v_{i}, e^{p}_{\ell})h_{p}(v_{j}, e^{p}_{\ell}) = a_{ij},
\label{eq:pairwise_hypergraph_similarity} \vspace{-0.1cm}
\end{equation}
where $\eta_{p}(e^{p}_{\ell})$ is the corresponding hyperedge weight of
$e^{p}_{\ell}=(v_{m}, v_{n})$ such that $\eta_{p}(e^{p}_{\ell})=a_{mn}$.
As a result, we have a weighted hypergraph
similarity matrix $U=(u_{ij})_{N\times N}$ that characterizes
the pairwise affinity relationships between vertices.
For simplicity, the resulting pairwise hypergraph similarity matrix is represented
as its corresponding matrix form: \vspace{-0.1cm}
\begin{equation}
U = H_{p}\Sigma_{p}H_{p}^{T} = A, \vspace{-0.1cm}
\end{equation}
where
$\Sigma_{p}$ is a diagonal matrix whose diagonal elements are denoted as
$(\eta_{p}(e^{p}_{\ell}))_{e^{p}_{\ell}\in E_{p}}$.

\begin{figure}[t]
\vspace{-0.2cm}
\begin{center}
\includegraphics[width=0.98\linewidth]{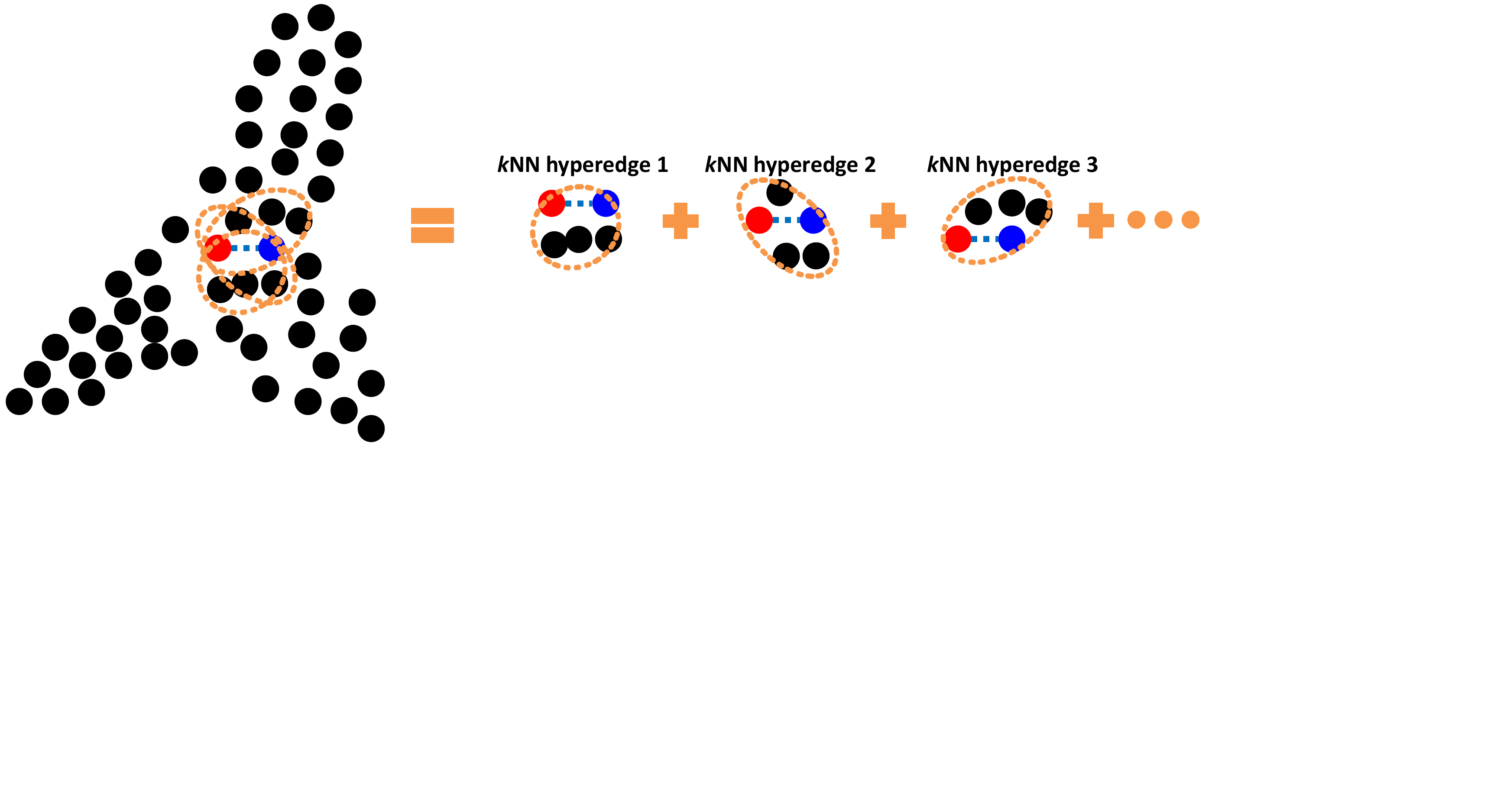}
\end{center} \vspace{-0.36cm}
\caption{Intuitive illustration of \KNN hypergraph construction. The left part shows a set of samples while the right part
displays a collection of \KNN hyperedges containing the two highlighted samples. Clearly, the \KNN hyperedges encode
the local neighboring information among data samples.}
\label{fig:knn_hypergraph} \vspace{-0.38cm}
\end{figure}

\emph{2) $k$-nearest-neighbor (\KNN) hypergraph.}
Based on the neighboring information on vertices, we further define a
\KNN hypergraph $G_{n}$, as shown in Fig.~\ref{fig:hypergraph_demo}.
For each vertex $v_{\ell}$, we search its corresponding $k$
nearest neighbors $\{v_{q}|v_{q}\in \mathcal{N}_{v_{\ell}}^{k}\}$ (as shown in Fig.~\ref{fig:clustering_example}(a)), and then use these nearest neighbors to form
a \KNN hyperedge $e_{\ell}^{n}$.
By concatenating all the \KNN hyperedges, a  \KNN hyperedge set is generated
as $E_{n}=\{e_{\ell}^{n}\}_{\ell=1}^{N}$, as illustrated in Fig.~\ref{fig:knn_hypergraph}.
To characterize the vertex-to-hyperedge membership, we define an indicator
function as: \vspace{-0.13cm}
\begin{equation}
\mathbb{I}(v_{i}, e^{n}_{\ell})=
\left\{
\begin{array}{cc}
1, & \mbox{if} \thickspace v_{i}\in e^{n}_{\ell},\\
0, & \mbox{otherwise},
\end{array}
\right. \vspace{-0.13cm}
\label{eq:indicator_function}
\end{equation}
Based on this indicator function,
we design a  hypergraph
model for softly assigning a vertex to each hyperedge: \vspace{-0.13cm}
\begin{equation}
h_{n}(v_{i}, e^{n}_{\ell}) =
\frac{a_{\ell i}\sqrt{\mathbb{I}(v_{i}, e^{n}_{\ell})}}{\sqrt{\sum_{t=1}^{N}\delta_{t}\mathbb{I}(v_{i}, e^{n}_{t})a_{ti}^{2}}},
\label{eq:KNN_hypergraph_incidence_matrix} \vspace{-0.13cm}
\end{equation}
where $\delta_{t}$ is the hyperedge weight associated with the $t$-th hyperedge $e^{n}_{t}$ such that
$\delta_{t}=\frac{1}{|e^{n}_{t}|}\sum_{j\in \{r|v_{r}\in e^{n}_{t}\}}a_{tj}$,
and $a_{\ell i}$ represents the vertex-to-hyperedge similarity between $v_{i}$ and the $\ell$-th $k$NN hyperedge $e_{\ell}^{n}$.
Specifically, $e_{\ell}^{n}$ is composed of a centroid vertex $v_{\ell}$
and its corresponding $k$ nearest vertices. Based on Eq.~\eqref{eq:pairwise_weight},
the vertex-to-hyperedge similarity $a_{\ell i}$
is computed as the pairwise similarity between $v_{i}$ and $v_{\ell}$.
As a result, we obtain a \KNN hypergraph incidence matrix $H_{n}=(h_{n}(v_{i}, e^{n}_{\ell}))_{|\mathbb{V}|\times |E_{n}|}$
for capturing the vertex-to-hyperedge relationships.
Based on $H_{n}$, a \KNN hypergraph similarity $b_{ij}$ between $v_{i}$ and $v_{j}$ is derived as: \vspace{-0.1cm}
\begin{equation}
\begin{array}{ll}
b_{ij} &= \underset{e^{n}_{\ell}\in E_{n}}{\sum}\delta_{\ell}\hspace{0.05cm} h_{n}(v_{i}, e^{n}_{\ell})h_{n}(v_{j}, e^{n}_{\ell})\\
&=
\frac{\sum_{\ell=1}^{N}(a_{\ell i}\sqrt{\mathbb{I}(v_{i}, e^{n}_{\ell})\delta_{\ell}})(a_{\ell j}\sqrt{\mathbb{I}(v_{j}, e^{n}_{\ell})\delta_{\ell}})}
{\sqrt{\sum_{t=1}^{N}a_{ti}^{2}(\sqrt{\mathbb{I}(v_{i}, e^{n}_{t})\delta_{t}})^{2}}
\sqrt{\sum_{t=1}^{N}a_{tj}^{2}(\sqrt{\mathbb{I}(v_{j}, e^{n}_{t})\delta_{t}})^{2}}}\\
&
=\frac{ \left<\mathbf{x}_{i}, \mathbf{x}_{j} \right>}
{\|\mathbf{x}_{i}\|\|\mathbf{x}_{j}\|},
\end{array}
\label{eq:KNN_hypergraph_similarity_matrix} \vspace{-0.1cm}
\end{equation}
where
$\mathbf{x}_{m}=\left(a_{\ell m}\sqrt{\mathbb{I}(v_{m}, e^{n}_{\ell})\delta_{\ell}}\right)_{\ell=1}^{N}$,
$<\hspace{-0.08cm}\cdot,\cdot\hspace{-0.08cm}>$ is the inner product operator,
and $\|\hspace{-0.05cm}\cdot\hspace{-0.05cm}\|$ is the 2-norm.
Indeed, $\mathbf{x}_{m}$ is a vertex-to-hyperedge feature vector
that characterizes the correlation between $v_{m}$ and the $k$NN hyperedges.
For example, the $\ell$-th element of $\mathbf{x}_{m}$ contains two terms: $a_{\ell m}$ and $\sqrt{\mathbb{I}(v_{m}, e^{n}_{\ell})\delta_{\ell}}$.
The first term $a_{\ell m}$ is the pairwise similarity between $v_{m}$ and the  vertex $v_{\ell}$ of the $\ell$-th
$k$NN hyperedge, and the second term $\sqrt{\mathbb{I}(v_{m}, e^{n}_{\ell})\delta_{\ell}}$
measures the cohesiveness of the $\ell$-th
$k$NN hyperedge by computing the average connection
similarity between the  vertex $v_{\ell}$
and the other vertices in the $\ell$-th
$k$NN hyperedge.
Thus, the \KNN hypergraph similarity $b_{ij}$ can be interpreted as the cosine
similarity between two vertex-to-hyperedge feature vectors $\mathbf{x}_{i}$ and $\mathbf{x}_{j}$.
As a result, we obtain a weighted similarity matrix $B=(b_{ij})_{N\times N}$ associated with $G_{n}$.
Essentially, $B=(b_{ij})_{N\times N}$ aims to explore the local neighboring relationships between
vertices. For simplicity, the \KNN hypergraph similarity matrix $B$ is denoted
as its corresponding matrix form: $B=H_{n}\Sigma_{n}H_{n}^{T}$ where $\Sigma_{n}$
is a diagonal matrix with the $\ell$-th diagonal element being $\delta_{\ell}$.
Fig.~\ref{fig:hypergraph_demo} gives an illustration of the process of the \KNN hypergraph construction.

\begin{figure}[t]
\vspace{-0.2cm}
\begin{center}
\includegraphics[width=0.98\linewidth]{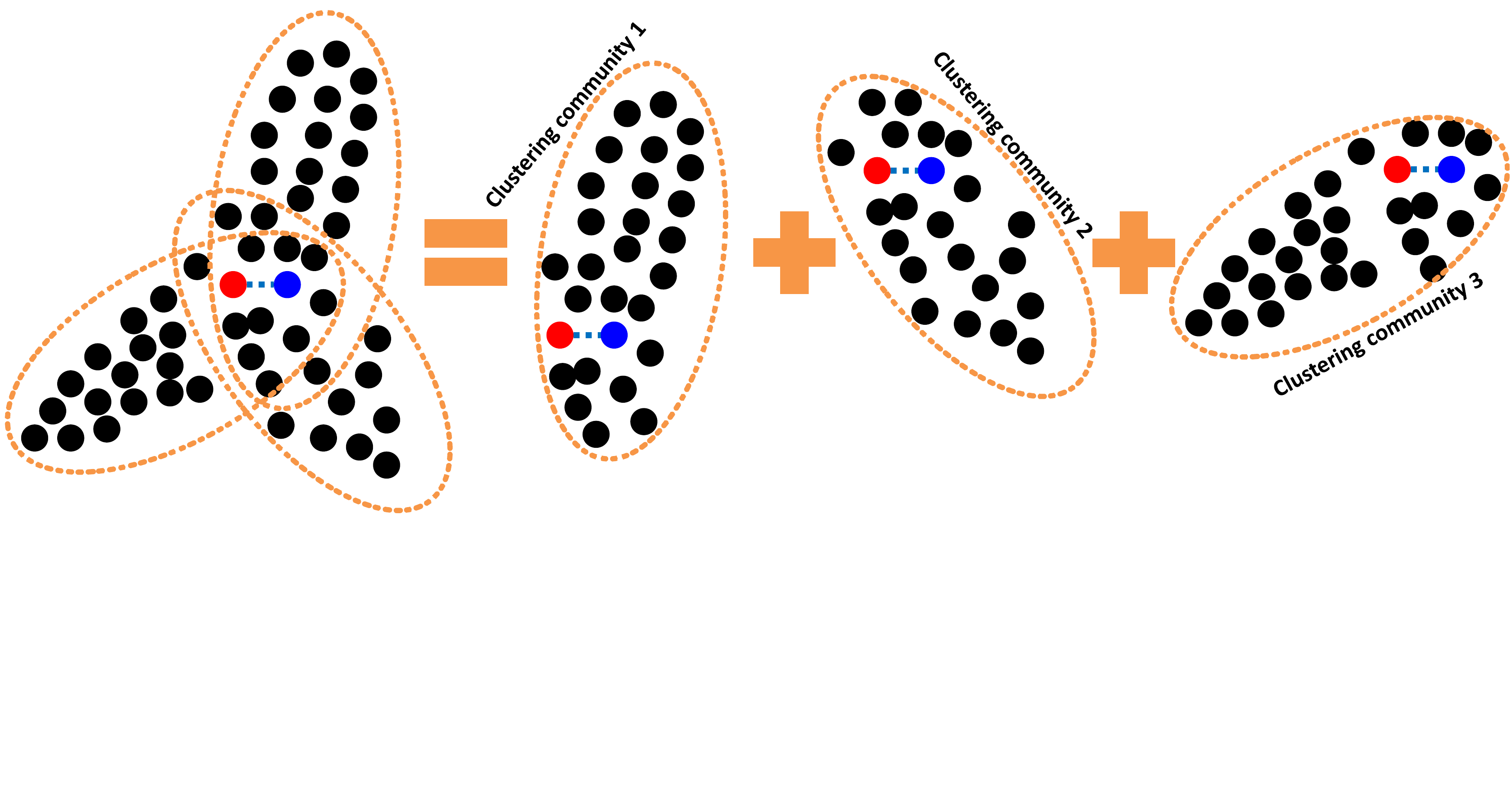}
\end{center} \vspace{-0.36cm}
\caption{Intuitive illustration of high-order over-clustering hypergraph construction. The left part shows a set of samples while the right part
displays a collection of high-order clustering hyperedges containing the two highlighted samples. Clearly, the high-order clustering hyperedges
encodes the manifold structure information among data samples at larger scales.}
\label{fig:clustering_hypergraph} \vspace{-0.48cm}
\end{figure}

\emph{3) High-order over-clustering hypergraph.}
In practice, the vertices are often distributed in different cohesive
vertex communities, and each community contains a set of mutually
correlated vertices with some common properties.  In order to
effectively discover such cohesive vertex communities, we propose a
high-order over-clustering hypergraph $G_{o}$ based on over-clustering
(or over-segmentation) using different clustering methods. Specifically, an over-clustering mechanism is
employed to generate a set of vertex groups, each of which corresponds
to a cohesive vertex community (as shown in
Fig.~\ref{fig:clustering_example}(a)). In this case, the vertices
belonging to the same vertex community are mutually influenced, and
work as the high-order contexts of the other vertices in the same
vertex community, as illustrated in Fig.~\ref{fig:clustering_hypergraph}. Without loss of generality, we assume that there are
$\mathcal{L}$ vertex communities in total.  For
convenience, let $E_{o} = \{e^{o}_{\ell}\}_{\ell=1}^{\mathcal{L}}$
denote these vertex communities, each of which corresponds to a
hyperedge $e_{\ell}^{o}$.  Based on these hyperedges, we define the
high-order over-clustering hypergraph incidence matrix
$H_{o}=(h_{o}(v_{i}, e^{o}_{\ell}))_{|\mathbb{V}|\times \mathcal{L}}$ as: \vspace{-0.1cm}
\begin{equation}
h_{o}(v_{i}, e^{o}_{\ell}) =
\frac{\sqrt{\mathbb{I}(v_{i}, e_{\ell}^{o})(1 + \frac{1}{|\mathfrak{N}_{i}^{\ell}|}\sum_{m \in \mathfrak{N}_{i}^{\ell}}a_{mi})}}
{\sqrt{\underset{e^{o}_{\ell}\in E_{o}}{\sum}\mu_{\ell}\mathbb{I}(v_{i}, e_{\ell}^{o})(1 + \frac{1}{|\mathfrak{N}_{i}^{\ell}|}\sum_{m \in \mathfrak{N}_{i}^{\ell}}a_{mi})}},
\label{eq:over-clustering_hypergraph_incidence_matrix} \vspace{-0.1cm}
\end{equation}
where $\mathbb{I}(\cdot, \cdot)$ is the indicator function in Eq.~\eqref{eq:indicator_function},
$\mathfrak{N}_{i}^{\ell}$ is the corresponding vertex index set of the nearest neighbors of $v_{i}$ in the hyperedge
$e^{o}_{\ell}$ (s.t. $|\mathfrak{N}_{i}^{\ell}|=3$ in the experiments), and $\mu_{\ell}$ is the associated
hyperedge weight of $e_{\ell}^{o}$ such that: \vspace{-0.1cm}
\begin{equation}
\mu_{\ell}=\frac{1}{2}\left(1 + \frac{1}{|e^{o}_{\ell}|}\sum_{i \in \{q| v_{q}\in e^{o}_{\ell}\}}\sum_{m \in \mathfrak{N}_{i}^{\ell}}\frac{a_{mi}}{|\mathfrak{N}_{i}^{\ell}|}\right). \vspace{-0.1cm}
\label{eq:over_clustering_mu}
\end{equation}
Here, $a_{mi}$ is the vertex-to-hyperedge similarity between $v_{i}$ and the $\ell$-th over-clustering
hyperedge $e^{o}_{\ell}$. To ensure the robustness of similarity evaluation, we only take into account
the affinity relationships between $v_{i}$ and its corresponding nearest neighbors (indexed by $\mathfrak{N}_{i}^{\ell}$) in $e^{o}_{\ell}$.
Therefore, $a_{mi}$ is the similarity between $v_{i}$ and the $m$-th vertex of $\mathfrak{N}_{i}^{\ell}$.
With the definition of $H_{o}$, the high-order over-clustering
hypergraph similarity $c_{ij}$ between $v_{i}$ and $v_{j}$ is
formulated as: \vspace{-0.1cm}
\begin{equation}
\hspace{-0.26cm}
\begin{array}{ll}
c_{ij} \hspace{-0.0cm} = \hspace{-0.1cm}\underset{e^{o}_{\ell}\in
E_{o}}{\sum}\mu_{\ell}\hspace{0.05cm}h_{o}(v_{i},
e^{o}_{\ell})h_{o}(v_{j}, e^{o}_{\ell}) &\vspace{0.1cm}\\
 = \hspace{-0.1cm}
\frac{
\underset{e^{o}_{\ell}\in E_{o}}{\sum}
\sqrt{\mu_{\ell}\mathbb{I}(v_{i}, e_{\ell}^{o})\left(1 + \frac{\sum_{m \in \mathfrak{N}_{i}^{\ell}}a_{mi}}{|\mathfrak{N}_{i}^{\ell}|}\right)}
\sqrt{\mu_{\ell}\mathbb{I}(v_{j}, e_{\ell}^{o})\left(1 + \frac{\sum_{m \in \mathfrak{N}_{j}^{\ell}}a_{mj}}{|\mathfrak{N}_{j}^{\ell}|}\right)}
}
{
\sqrt{\underset{e^{o}_{\ell}\in E_{o}}{\sum}\mu_{\ell}\mathbb{I}(v_{i}, e_{\ell}^{o})\left(1 + \frac{\sum_{m \in \mathfrak{N}_{i}^{\ell}}a_{mi}}{|\mathfrak{N}_{i}^{\ell}|}\right)}
\sqrt{\underset{e^{o}_{\ell}\in E_{o}}{\sum}\mu_{\ell}\mathbb{I}(v_{j}, e_{\ell}^{o})\left(1 + \frac{\sum_{m \in \mathfrak{N}_{j}^{\ell}}a_{mj}}{|\mathfrak{N}_{j}^{\ell}|}\right)}
} &
\vspace{0.1cm}\\
 = \hspace{-0.1cm}\frac{ \left<\mathbf{y}_{i},
\mathbf{y}_{j}  \right>}{\|\mathbf{y}_{i}\|\|\mathbf{y}_{j}\|}, &
\end{array}
\hspace{-0.2cm}
\label{eq:over-clustering_hypergraph_similarity_matrix} \vspace{-0.1cm}
\end{equation}
where $\mathbf{y}_{q}$ is an $\mathcal{L}$-dimensional vector with the $\ell$-th element
being $\sqrt{\mu_{\ell}\mathbb{I}(v_{q}, e_{\ell}^{o})(1 + \frac{1}{|\mathfrak{N}_{q}^{\ell}|}\sum_{m \in \mathfrak{N}_{q}^{\ell}}a_{mq})}$.
Actually, $\mathbf{y}_{q}$ is a
vertex-to-hyperedge feature vector,
and its $\ell$-th element
consists of two components: $\mu_{\ell}$ and $\mathbb{I}(v_{q}, e_{\ell}^{o})(1 + \frac{1}{|\mathfrak{N}_{q}^{\ell}|}\sum_{m \in \mathfrak{N}_{q}^{\ell}}a_{mq})$.
As defined in Eq.~\eqref{eq:over_clustering_mu}, the left term measures the average cross-link degree of the within-community-$\ell$ vertices
with respect to the other vertices in the same community, while the right term
reflects the average affinity relationships between $v_{q}$ and
the vertices in the $\ell$-th community.
Therefore, the high-order over-clustering hypergraph similarity $c_{ij}$ can be viewed as
the cosine similarity between two vertex-to-hyperedge feature vectors $\mathbf{y}_{i}$ and $\mathbf{y}_{j}$.
As a result, a weighted similarity matrix $C=(c_{ij})_{N\times N}$ is obtained to capture the local grouping
information on vertices. For simplicity, the high-order over-clustering hypergraph similarity matrix $C$ can be expressed
as its corresponding matrix form: $C=H_{o}\Sigma_{o}H_{o}^{T}$, where $\Sigma_{o}$
is a diagonal matrix with the $\ell$-th diagonal element being $\mu_{\ell}$.

\begin{figure}[t]
\vspace{-0.2cm}
\begin{center}
\includegraphics[width=0.46\linewidth]{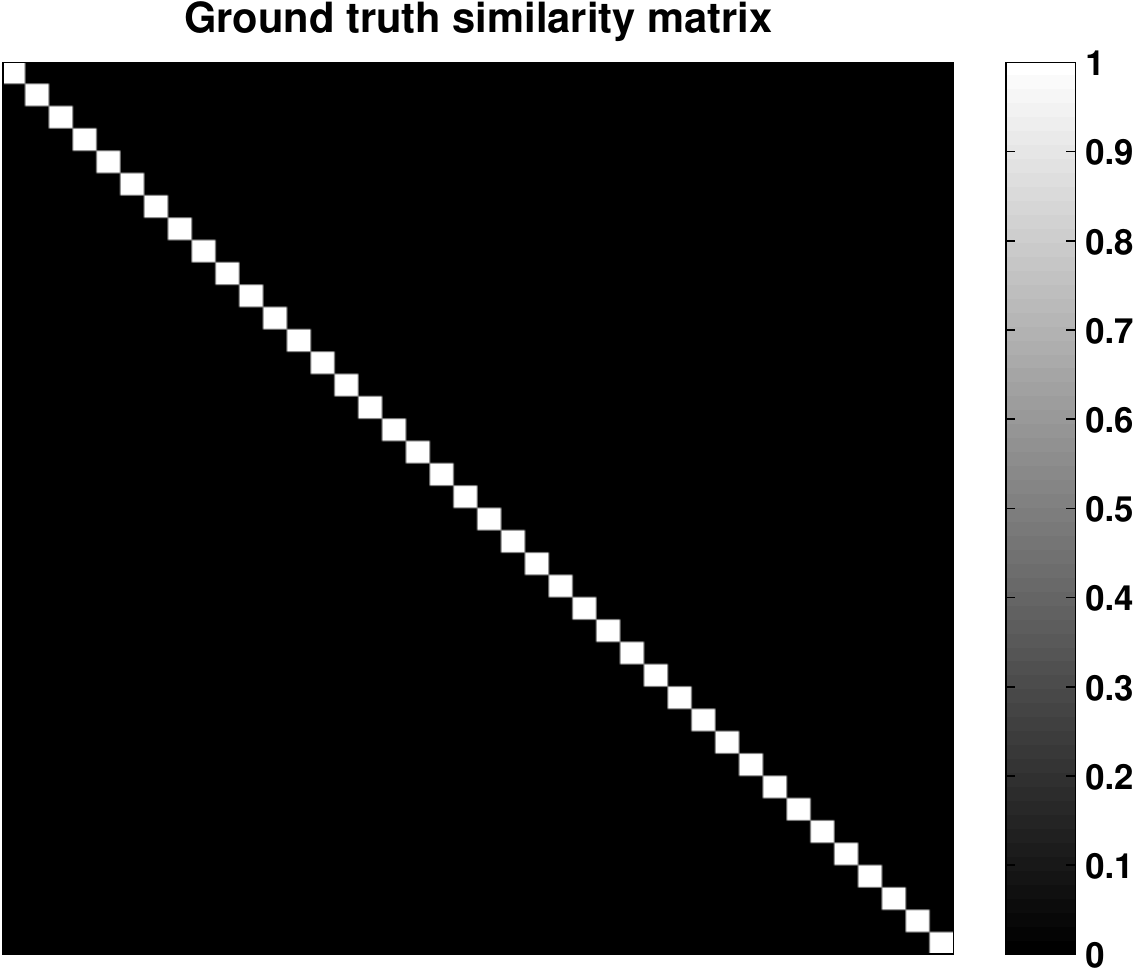}\hspace{0.1cm}
\includegraphics[width=0.46\linewidth]{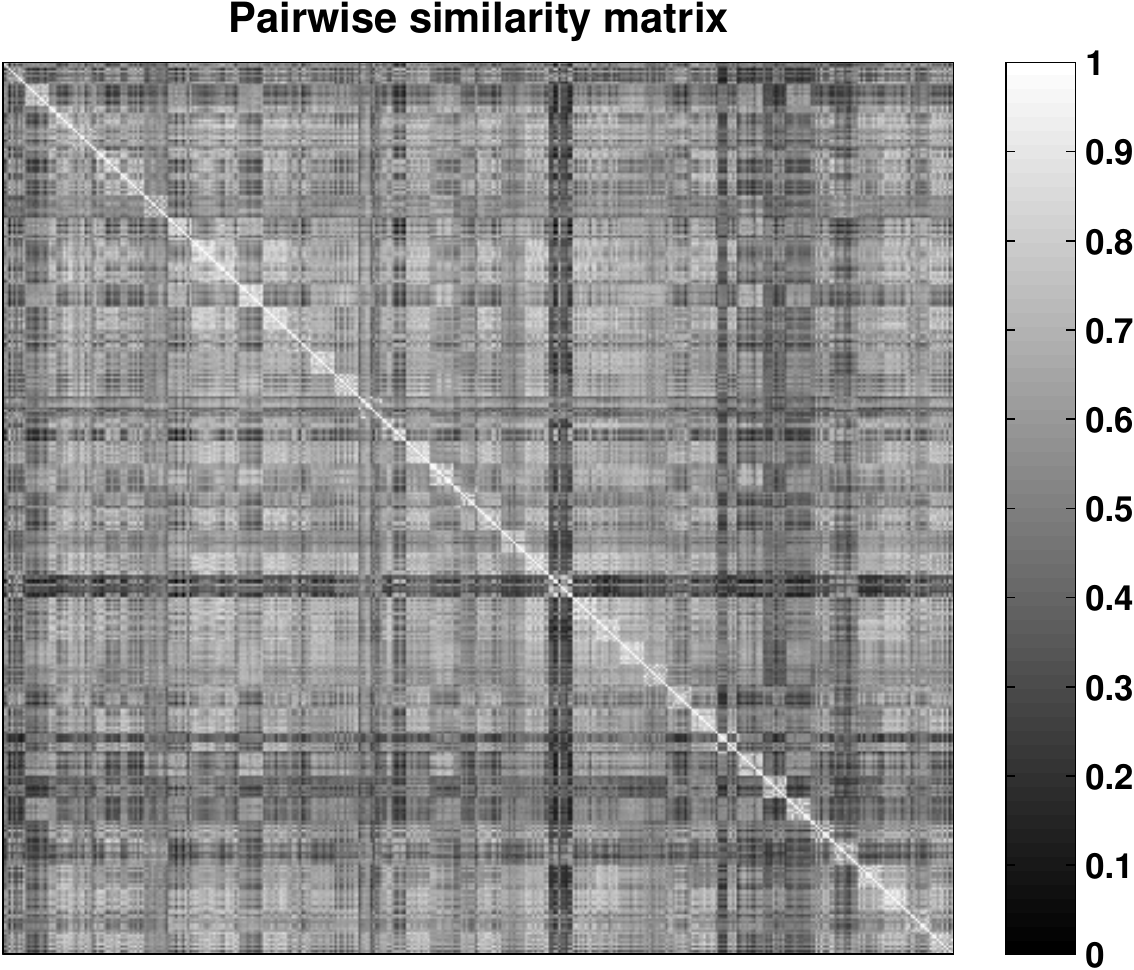}\\
\includegraphics[width=0.46\linewidth]{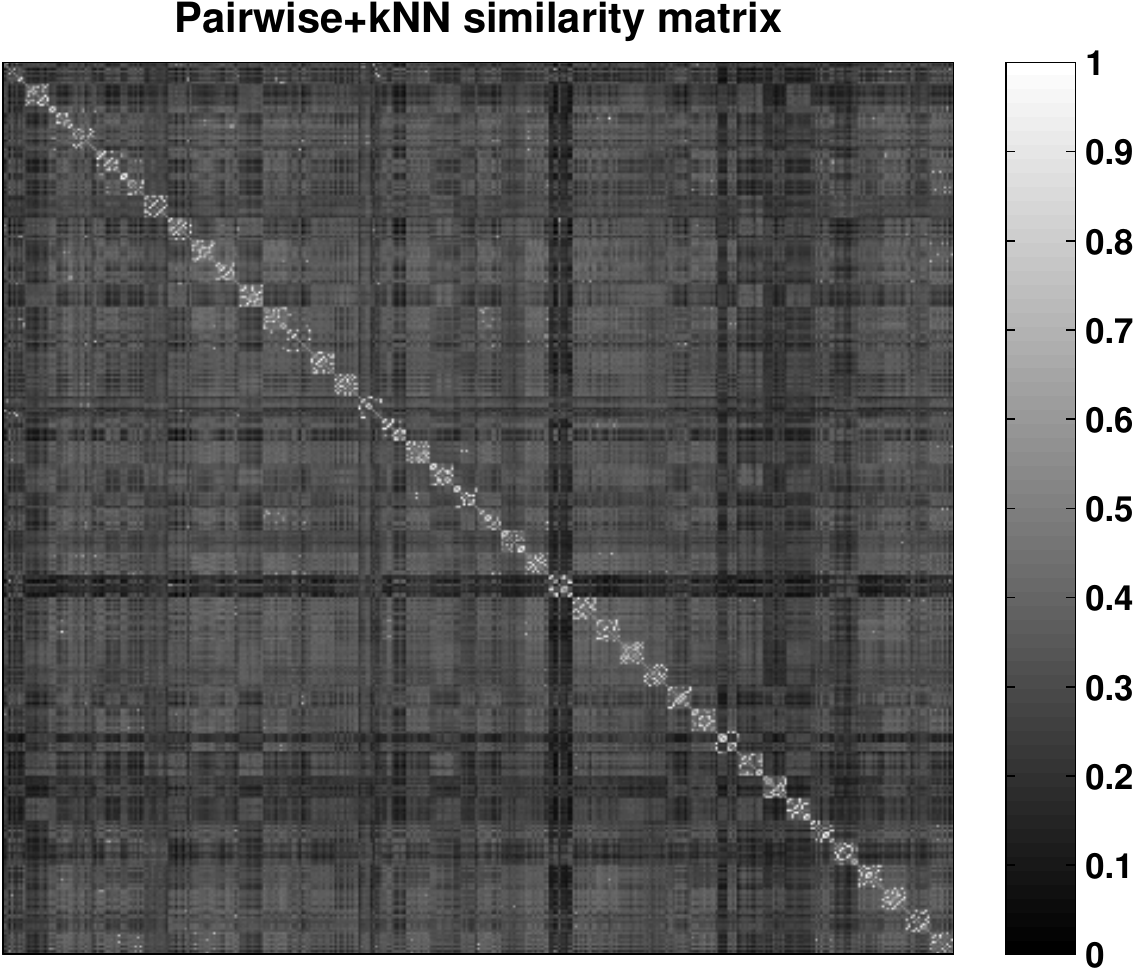} \hspace{0.1cm}
\includegraphics[width=0.46\linewidth]{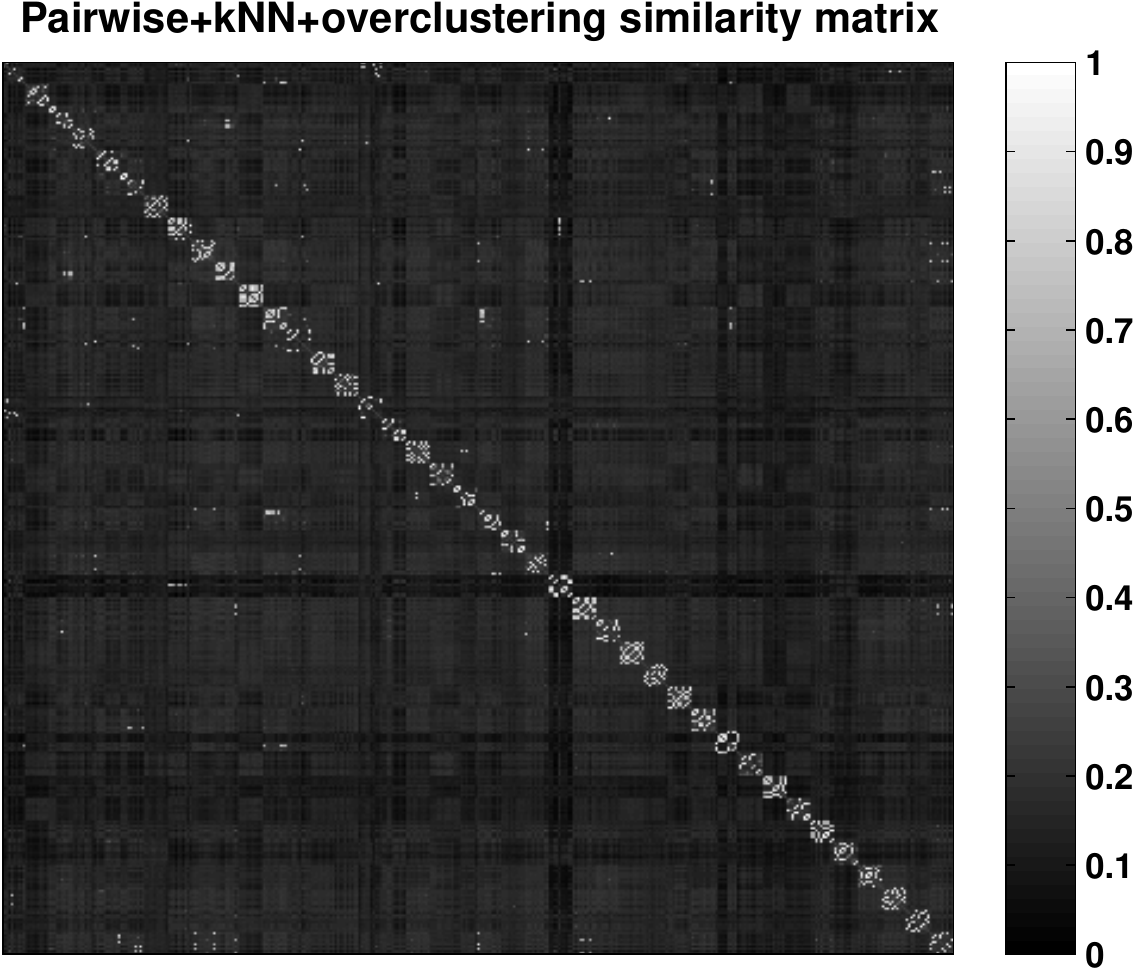}
\end{center}
\vspace{-0.26cm}
\caption{Illustration of different types of hypergraph similarity matrices. Specifically, the top-left subfigure shows the ground truth similarity matrix
of the ORL face dataset (referred to in Sec.~\ref{sec:Experiment}); the top-right subfigure displays the pairwise hypergraph similarity matrix;
the bottom-left subfigure exhibits the pairwise+\KNN hypergraph similarity matrix; the bottom-right subfigure draws
our pairwise+\KNN+over-clustering hypergraph similarity matrix. Clearly, our hypergraph similarity matrix is the closest to
the ground truth.}
\label{fig:similarity_matrix}
\vspace{-0.46cm}
\end{figure}

Fig.~\ref{fig:similarity_matrix} gives an example of showing the different hypergraph similarity matrices on
the ORL face dataset (referred to in Sec.~\ref{sec:Experiment}).
By linearly combining the above three types of hypergraphs, a context-aware hypergraph similarity matrix $S=(s_{ij})_{N\times N}$ is obtained as follows: \vspace{-0.1cm}
\begin{equation}
S = \mathcal{H}
\left(
\begin{array}{ccc}
\alpha\Sigma_{p} & \mathbf{0} & \mathbf{0}\\
\mathbf{0} &\beta\Sigma_{n}  & \mathbf{0}\\
\mathbf{0} & \mathbf{0} &(1-\alpha-\beta)\Sigma_{o} \\
\end{array}
\right)\mathcal{H}^{T},
\label{eq:hyper_similarity_measure} \vspace{-0.1cm}
\end{equation}
where $\mathcal{H}=(H_{p}\thinspace H_{n}\thinspace H_{o})$ and
$(\alpha, \beta)$ are the nonnegative weighting factors such that
$(\alpha+\beta)\leq 1$.
Encoding the local neighboring information,
the \KNN hypergraph plays a role in locally smoothing
the clustering results (obtained by only using the pairwise hypergraph), as shown in the bottom-left
part of Fig.~\ref{fig:similarity_matrix}.
By constructing the high-order
over-clustering hypergraph,
we are capable of capturing the
manifold structure information among data samples at larger scales,
which leads to more accurate clustering results (shown in the bottom-right
part of Fig.~\ref{fig:similarity_matrix}).
Therefore, the final context-aware hypergraph similarity matrix
keeps a balance among the three types of hypergraph information.
In practice, it is easy to emphasize one
particular type of hypergraph information by enlarging its associated weight.
Fig.~\ref{fig:clustering_example} (b) gives an example of illustrating the
hypergraph clustering results based on
the above hypergraph similarities.

\vspace{-0.18cm}
\subsection{Discriminative hypergraph partitioning for spectral clustering \label{sec:final_hypergraph_partitioning}}

Having defined a context aware vertex similarity measure, we now propose
a discriminative hypergraph partitioning criterion based on this measure,
with its corresponding optimization procedure.

\emph{1) Preliminaries of hypergraph partitioning.}
Hypergraph partitioning seeks an optimal hypergraph cut solution
for effective data clustering. $\mathcal{K}$-way normalized cut~\cite{Yu-Shi14} is a well-known hypergraph partitioning
criterion, which aims to optimally
partition the vertex set $\mathbb{V}$ into $\mathcal{K}$ disjoint subsets (i.e.,
$\mathbb{V}=\bigcup_{l=1}^{\mathcal{K}}\mathbb{V}_{l}$ s.t.
$\mathbb{V}_{m}\bigcap \mathbb{V}_{n}=\emptyset$, $\forall m\neq n$)
by solving the following optimization problem: \vspace{-0.12cm}
\begin{equation}
\begin{array}{cc}
\mbox{\bf{max}} & f(X)=\frac{1}{\mathcal{K}}\sum_{n=1}^{\mathcal{K}}\frac{X_{n}^{T}S
X_{n}}{X_{n}^{T}D X_{n}},
\vspace{0.1cm} \\  \mbox{\bf{s.t.}} & X
\in \{0,1\}^{N \times \mathcal{K}}, X\mathbbold{1}_{\mathcal{K}}=\mathbbold{1}_{N},
\end{array}
\label{eq:GraphOptimization1} \vspace{-0.12cm}
\end{equation}
where $X$ is an $N \times \mathcal{K}$ partition matrix
such that $X^{T}X$ is a diagonal matrix,
$\mathbbold{1}_{d}$
denotes a $d \times 1$ vector with each element being 1,
 $D$ is an $N \times N$ diagonal matrix with the
$m$-th diagonal element being the sum of the elements belonging to
the $m$-th row of $S$ for $1 \leq m \leq N$, and $X_{n}$ is the
$n$-th column of $X$ for $1 \leq n \leq \mathcal{K}$.
As pointed out in~\cite{Yu-Shi14}, the optimization problem~\eqref{eq:GraphOptimization1} is
typically relaxed to: \vspace{-0.1cm}
\begin{equation}
\begin{array}{cl}
\mbox{\bf{max}} & h(Z) = \frac{1}{\mathcal{K}} tr(Z^{T}SZ),\vspace{0.1cm}\\
\mbox{\bf{s.t.}} &  Z^{T}DZ = I_{\mathcal{K}},
\end{array}
\label{eq:trace_normalized_cut} \vspace{-0.1cm}
\end{equation}
where $I_{\mathcal{K}}$ is a $\mathcal{K} \times \mathcal{K}$ identity matrix,
$tr(\cdot)$ denotes the trace of a matrix, and $Z = X(X^{T}DX)^{-\frac{1}{2}}$.
Eq.~\eqref{eq:trace_normalized_cut} is a trace maximization problem and
can be solved by generalized eigenvalue decomposition.
To simultaneously capture both intra-cluster compactness and the inter-cluster separability among
the vertices in a unified clustering framework,
we propose a discriminative hypergraph partitioning criterion
which can be formulated as a trace-ratio optimization problem.

\begin{algorithm}[t!]
\scriptsize \hspace{-0.1cm}
  \KwIn
  {
   A  dataset $\mathbb{Z}=\{\mathbf{z}_{i}\}_{i=1}^{N}$ and the number of clusters $\mathcal{K}$
 }

  \begin{enumerate} \itemindent=-6pt
  \item Obtain the hypergraph similarity matrix  $S=(s_{ij})_{N\times N}$.
        \begin{itemize}\itemindent=-9pt
           \item Compute the pairwise hypergraph similarity $u^{p}_{ij}$ in Eq.~\eqref{eq:pairwise_hypergraph_similarity}.
           \item Obtain the \KNN hypergraph similarity $u^{n}_{ij}$ in Eq.~\eqref{eq:KNN_hypergraph_similarity_matrix}.
           \item Compute the high-order over-clustering hypergraph \\similarity $u^{c}_{ij}$ in Eq.~\eqref{eq:over-clustering_hypergraph_similarity_matrix}.
           \item Combine the above three similarities to generate $s_{ij}$ by Eq.~\eqref{eq:hyper_similarity_measure}.
        \end{itemize}
  \item Perform spectral graph partitioning.
  \begin{itemize}\itemindent=-9pt
      \item  Compute the graph Laplacian matrix $Q=D-S$.
      \item  Solve the optimization problem~\eqref{eq:GraphOptimization5} by the Newton-Lanczos algorithm.
      \item  Calculate a candidate graph partitioning solution $\widetilde{X}$  by:
            $\widetilde{X}=\mbox{Diag}(\mbox{diag}^{-\frac{1}{2}}({P}{P}^{T})){P}$.
      \item  Iteratively refine $\widetilde{X}$ to find an optimal discrete solution $X$.
  \end{itemize}
  \end{enumerate}
   \hspace{-0.1cm}
    \KwOut
  {
      The optimal graph partitioning solution $X$.
  }
  \caption{The proposed pairwise+$k$NN+over-clustering hypergraph spectral clustering algorithm (PKO+HSC).\vspace{-1.35cm} \label{alg:Framwork}}
\end{algorithm}

\emph{2) Discriminative hypergraph partitioning criterion (DHPC).}
The proposed DHPC
considers both
the inter-cluster separability and the intra-cluster compactness,
and thus aims to solve the following optimization
problem: \vspace{-0.1cm}
\begin{equation}
\hspace{-0.12cm}
\begin{array}{cl}
\mbox{\bf{max}} \hspace{-0.12cm}& g(X)=\frac{1}{\mathcal{K}}\sum_{n=1}^{\mathcal{K}}\frac{X_{n}^{T}
S X_{n}}{X_{n}^{T} Q X_{n}}
 \vspace{0.06cm}\\&\hspace{0.8cm}=
\frac{1}{\mathcal{K}}\sum_{n=1}^{\mathcal{K}}\frac{[X_{n}(X_{n}^{T}X_{n})^{-\frac{1}{2}}]^{T}S[X_{n}(X_{n}^{T}X_{n})^{-\frac{1}{2}}]}
{[X_{n}(X_{n}^{T}X_{n})^{-\frac{1}{2}}]^{T} Q
[X_{n}(X_{n}^{T}X_{n})^{-\frac{1}{2}}]}, \vspace{0.1cm}
\\  \mbox{\bf{s.t.}} \hspace{-0.12cm} & X \in \{0,1\}^{N \times \mathcal{K}},
X\mathbbold{1}_{\mathcal{K}}=\mathbbold{1}_{N},
\end{array}
\hspace{-0.56cm}
\label{eq:GraphOptimization} \vspace{-0.1cm}
\end{equation}
where
$Q=D-S$.
In the proposed DHPC,
the intra-cluster compactness and the inter-cluster separability are
respectively captured by $X_{n}^{T} S  X_{n}$ and $X_{n}^{T}Q
X_{n}$, which are formulated as:  \vspace{-0.1cm}
\begin{equation}
X_{n}^{T} S X_{n}=\sum_{i\in \mathbb{V}_{n}}\sum_{j \in
\mathbb{V}_{n}}s_{ij}, \hspace{0.2cm}
 X_{n}^{T}Q X_{n}=\sum_{i \in \mathbb{V}_{n}}\sum_{j \notin \mathbb{V}_{n}}
s_{ij}, \vspace{0.0cm} \label{eq:intra-inter} \vspace{-0.1cm}
\end{equation}
where $\mathbb{V}_{n}$ denotes the vertex set belonging to the $n$-th
cluster. The larger the value of $X_{n}^{T} S X_{n}$, the more
compact the intra-cluster samples. The smaller the value of
$X_{n}^{T}Q X_{n}$, the more separable the inter-cluster samples. As
a result, an optimal hypergraph partitioning solution is obtained
by maximizing $g(X)$ in Eq.~\eqref{eq:GraphOptimization}.
For simplicity, let $P_{n}$ denote the vertex-to-cluster membership
vector associated with the $n$-th cluster such that
$P_{n}=X_{n}(X_{n}^{T}X_{n})^{-\frac{1}{2}}$, and $P$ denote
the vertex-to-cluster membership matrix that is a concatenation of
all the vertex-to-cluster membership
vectors
such that
$P=(P_{1}\thinspace P_{2}\thinspace \ldots \thinspace P_{\mathcal{K}})
=X(X^{T}X)^{-\frac{1}{2}}$.
It can be shown that $P$ is an orthogonal matrix: \vspace{-0.1cm}
\begin{equation}
\begin{array}{cl}
P^{T}P&=[X(X^{T}X)^{-\frac{1}{2}}]^{T}[X(X^{T}X)^{-\frac{1}{2}}]\\
&=(X^{T}X)^{-\frac{1}{2}}(X^{T}X)(X^{T}X)^{-\frac{1}{2}}=
I_{\mathcal{K}},
\end{array} \vspace{-0.1cm}
\end{equation}
where $X^{T}X$ is a diagonal matrix.
According to the conclusion in~\cite{Yu-Shi14}, we obtain
$\widetilde{X}=\mbox{Diag}(\mbox{diag}^{-\frac{1}{2}}(PP^{T}))P$
that is
the corresponding inverse transform of $P=X(X^{T}X)^{-\frac{1}{2}}$.
Here, $\mbox{Diag}(\cdot)$ denotes a
diagonal matrix formed from its vector argument, and
$\mbox{diag}(\cdot)$ represents a
column vector
formed from the diagonal elements of its matrix argument.
Consequently, the optimization problem in Eq.~\eqref{eq:GraphOptimization}
can be rewritten as: \vspace{-0.1cm}
\begin{equation}
\hspace{-0.1cm}
\begin{array}{cl}
\mbox{\bf{max}} \hspace{-0.1cm}& g(X)=\frac{1}{\mathcal{K}}\sum_{n=1}^{\mathcal{K}}\frac{P_{n}^{T}
S P_{n}}{P_{n}^{T} Q P_{n}}=\frac{1}{\mathcal{K}}\sum_{n=1}^{\mathcal{K}}\frac{tr(P_{n}^{T}
S P_{n})}{tr(P_{n}^{T} Q P_{n})},
\\
\mbox{\bf{s.t.}} \hspace{-0.1cm} & P^{T}P=I_{\mathcal{K}}.
\end{array}
\hspace{-0.25cm}
\label{eq:GraphOptimization2} \vspace{-0.1cm}
\end{equation}
This is a trace-ratio-sum optimization problem, which is non-convex and difficult to solve~\cite{Wang-Shen-Zheng-Ren-ACCV2009}. Thus, we approximate the original optimization problem~\eqref{eq:GraphOptimization2}
using the following sum-trace-ratio optimization problem: \vspace{-0.1cm}
\begin{equation}
\begin{array}{cl}
\mbox{\bf{max}} & f(P)=\frac{1}{\mathcal{K}} \frac{\sum_{n=1}^{\mathcal{K}}tr(P_{n}^{T}
S P_{n})}{\sum_{n=1}^{\mathcal{K}}tr(P_{n}^{T} Q P_{n})}
= \frac{1}{\mathcal{K}} \frac{tr(\sum_{n=1}^{\mathcal{K}}P_{n}^{T}
S P_{n})}{tr(\sum_{n=1}^{\mathcal{K}}P_{n}^{T} Q P_{n})},
\\
\mbox{\bf{s.t.}} & P^{T}P=I_{\mathcal{K}}.
\end{array}
\label{eq:GraphOptimization3} \vspace{-0.1cm}
\end{equation}
Due to $tr(\sum_{n=1}^{\mathcal{K}}P_{n}^{T} S P_{n})=tr(P^{T}SP)$ and
$tr(\sum_{n=1}^{\mathcal{K}}P_{n}^{T} Q P_{n})=tr(P^{T}QP)$, the above optimization problem~\eqref{eq:GraphOptimization3}
can be reformulated as: \vspace{-0.06cm}
\begin{equation}
\begin{array}{cl}
\mbox{\bf{max}} & f(P)=\frac{1}{\mathcal{K}} \frac{tr(P^{T}SP)}{tr(P^{T}QP)},
\\
\mbox{\bf{s.t.}} & P^{T}P=I_{\mathcal{K}}.
\end{array}
\label{eq:GraphOptimization5} \vspace{-0.06cm}
\end{equation}
The trace-ratio optimization problem~\eqref{eq:GraphOptimization5} has been investigated
in~\cite{Wang-Yan-Xu-Tang-Huang-CVPR2007,Jia-Nie-Zhang-TNN2009,Ngo-SIAM2012-traceratio,shen2008supervised}.
In order to obtain an effective solution to Eq.~\eqref{eq:GraphOptimization5},
we therefore utilize the Newton-Lanczos algorithm~\cite{Ngo-SIAM2012-traceratio} for trace-ratio maximization.
The Newton-Lanczos algorithm includes the following three iterative steps: \vspace{-0.1cm}
\begin{itemize} \itemsep = -1pt
\item Compute the trace ratio $\rho = \frac{tr(P^{T}SP)}{tr(P^{T}QP)}$;
\item Run the Lanczos algorithm~\cite{Lanczos_wiki} to compute the $\mathcal{K}$ largest eigenvalues of $S-\rho Q$ as well as
their associated eigenvectors $(P_{1}\thinspace P_{2}\thinspace \ldots \thinspace P_{\mathcal{K}})\equiv P$;
\item Repeat the above two steps until convergence. \vspace{-0.1cm}
\end{itemize}
In practice, the initial solution $P$ is chosen as
the $\mathcal{K}$ principal
eigenvectors (i.e., corresponding to
the $\mathcal{K}$ largest eigenvalues) of the matrix $Q^{-1}S$.
If $Q$ is a
singular matrix, $Q^{-1}S$ is replaced with the matrix
$(Q+\epsilon I_{N})^{-1}S$, where $I_{N}$ is an $N \times N$
identity matrix and $\epsilon$ is a small positive constant
($\epsilon=10^{-6}$ in the experiments).

\begin{figure}[t]
\begin{center}
\includegraphics[width=0.93\linewidth]{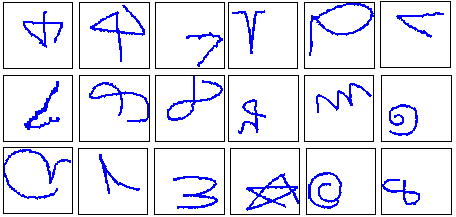}
\end{center}
\vspace{-0.28cm}
\caption{Representative samples of the trajectory dataset. Each subfigure is associated with a representative sample from
a particular trajectory cluster.} \vspace{-0.25cm}
\label{fig:traj_sample}
\end{figure}

\begin{figure}[t]
\begin{center}
\includegraphics[width=0.93\linewidth]{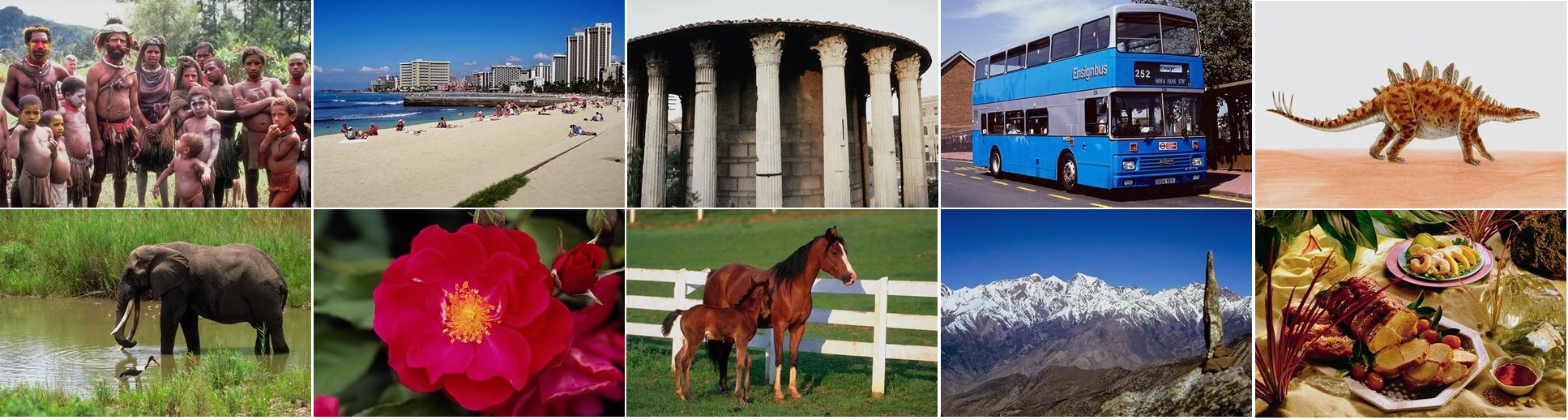}
\end{center}
\vspace{-0.25cm}
\caption{Representative samples of the Corel dataset. Each subfigure is associated with a representative sample from
a particular image cluster.} \vspace{-0.35cm}
\label{fig:Corel_sample}
\end{figure}

After solving the trace-ratio optimization problem~\eqref{eq:GraphOptimization5}, we obtain a candidate solution
$\widetilde{X}$ to Eq.~\eqref{eq:GraphOptimization} as follows: \vspace{-0.1cm}
\begin{equation}
\widetilde{X}=\mbox{Diag}(\mbox{diag}^{-\frac{1}{2}}(PP^{T}))P. \vspace{-0.1cm}
\end{equation}
However, the candidate solution $\widetilde{X}$ is a real-valued hypergraph
partitioning solution, and thus does not satisfy the discrete-solution
requirements for data clustering. As a result,
an iterative  refining procedure~\cite{Yu-Shi14} may be
used to find the optimal discrete hypergraph partitioning solution $X$ to Eq.~\eqref{eq:GraphOptimization}
(more details can be found in Steps four to eight of the algorithm in \cite{Yu-Shi14}).
After combining the constructed pairwise+$k$NN+over-clustering hypergraphs (referred to in Sec.~\ref{sec:Prob_hypergraph}), we have a DHPC-based
spectral clustering algorithm called PKO+HSC (pairwise+$k$NN+over-clustering hypergraph spectral clustering),
as listed in Algorithm~\ref{alg:Framwork}.

\begin{figure*}[t]
\begin{center}
\includegraphics[width=0.9\linewidth]{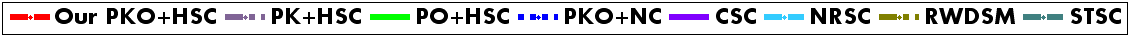}\\
\includegraphics[width=0.25\linewidth]{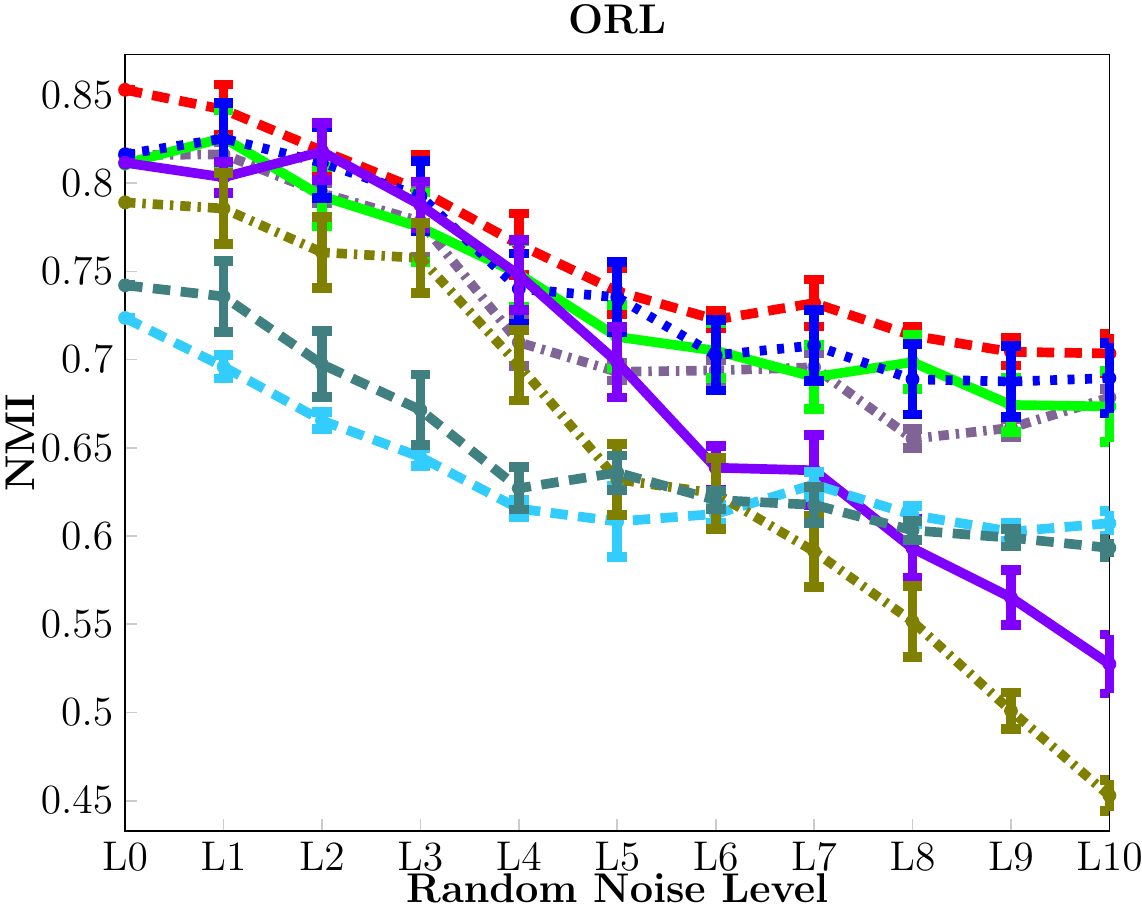} \hspace{-0.26cm}
\includegraphics[width=0.25\linewidth]{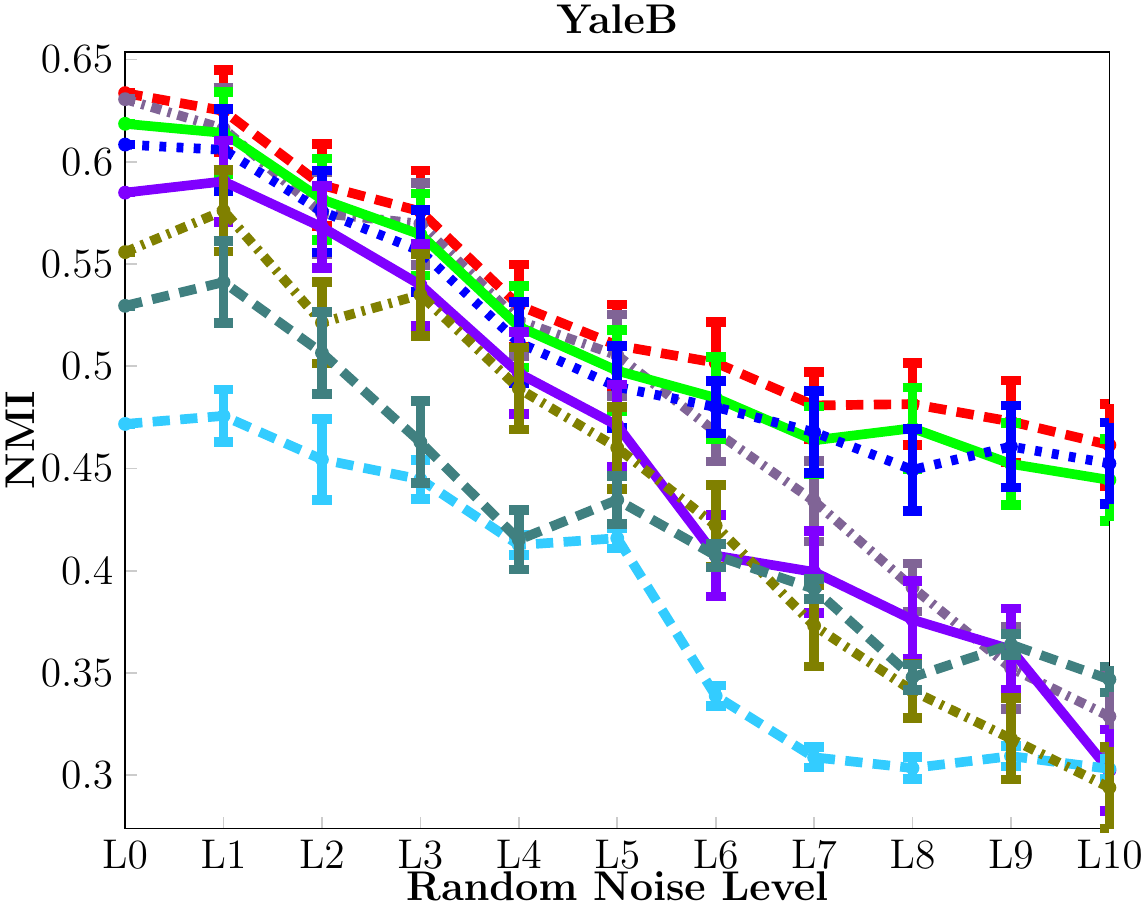} \hspace{-0.26cm}
\includegraphics[width=0.25\linewidth]{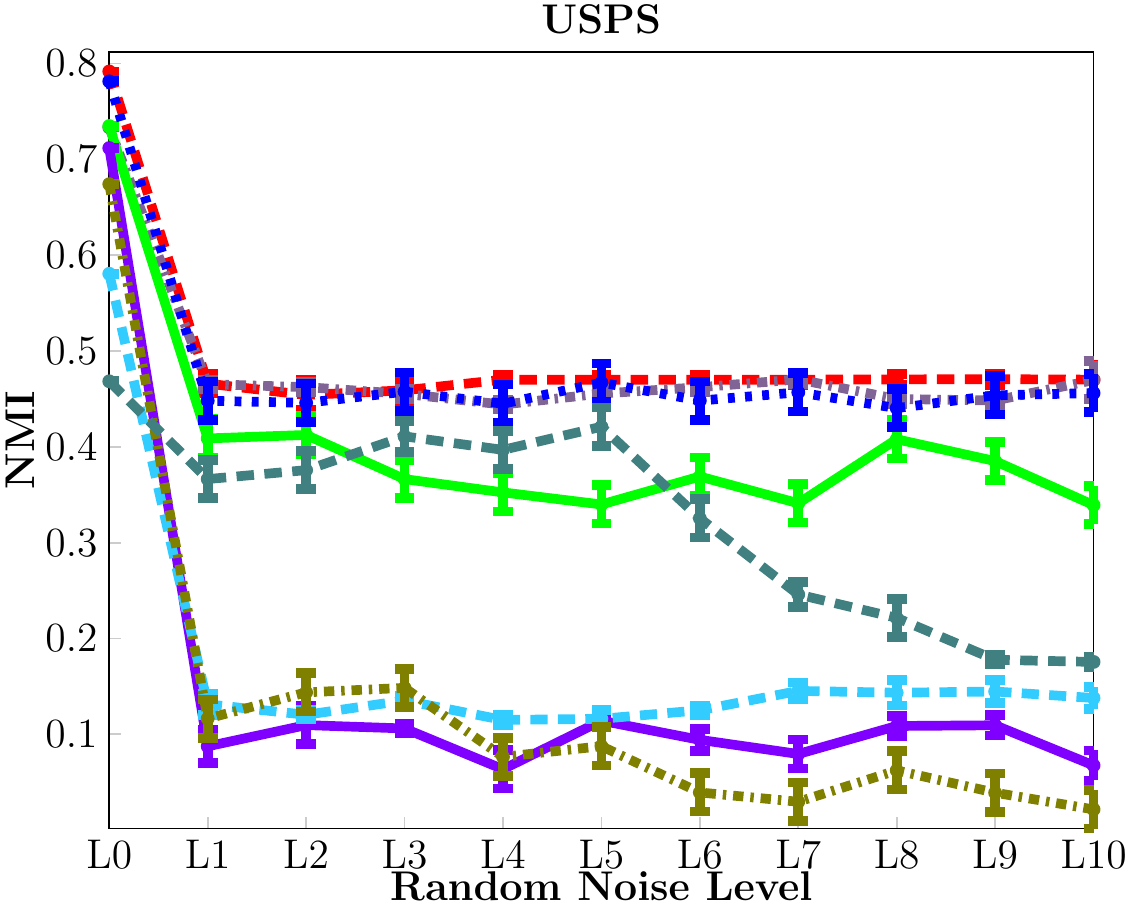} \hspace{-0.26cm}
\includegraphics[width=0.25\linewidth]{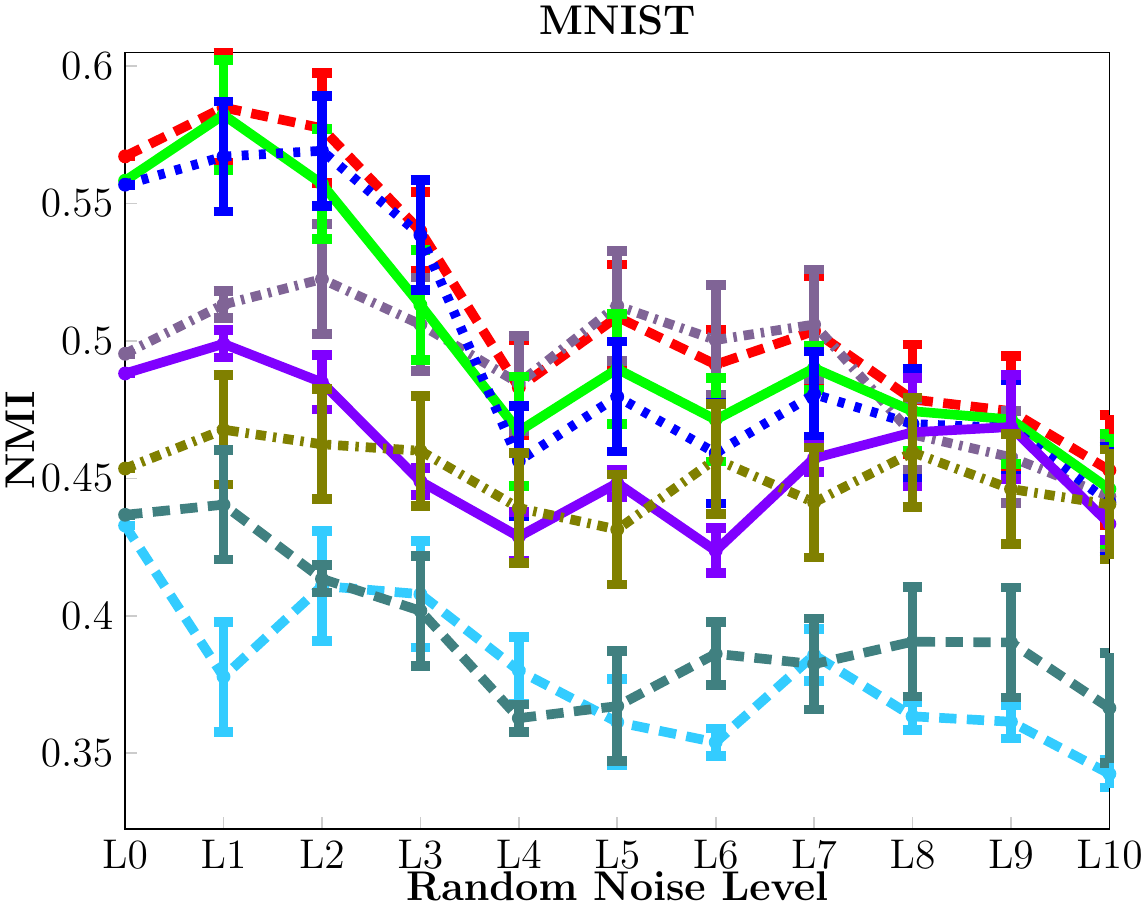}\\
\includegraphics[width=0.25\linewidth]{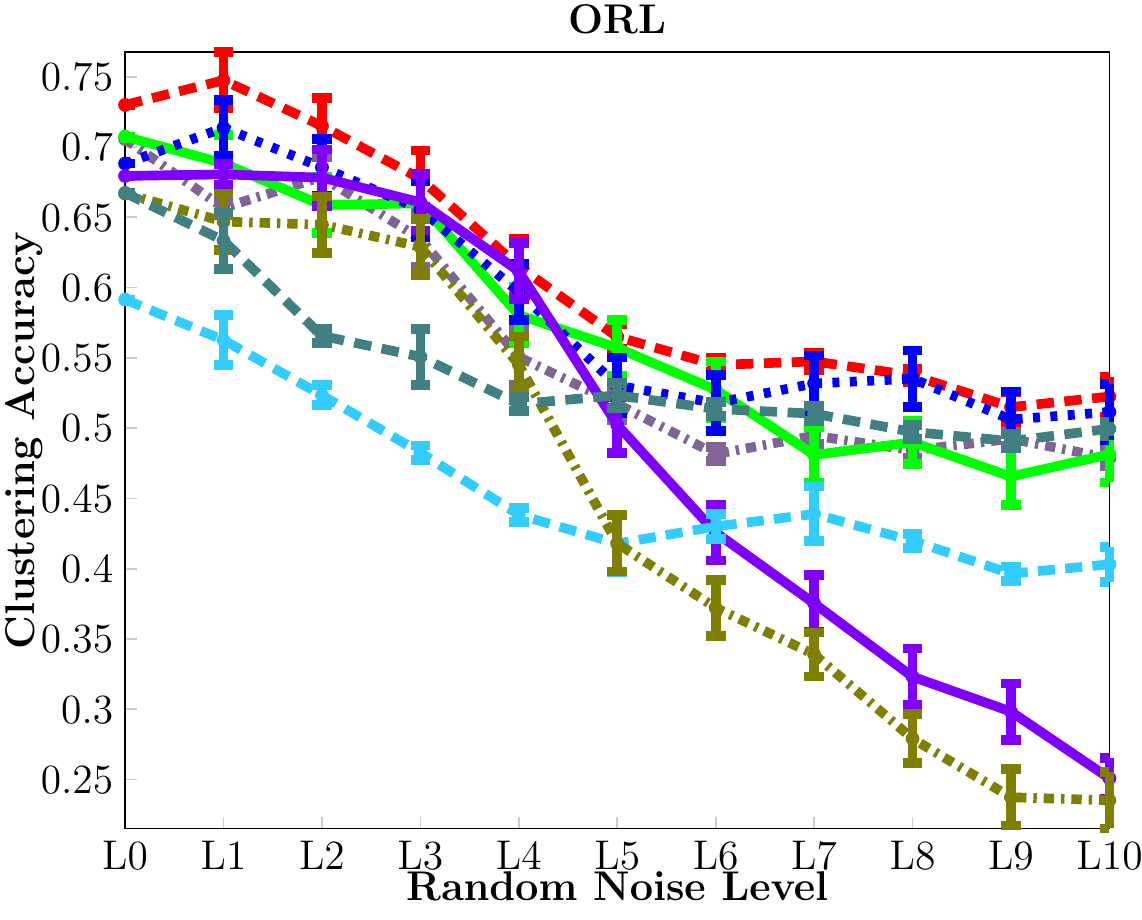} \hspace{-0.26cm}
\includegraphics[width=0.25\linewidth]{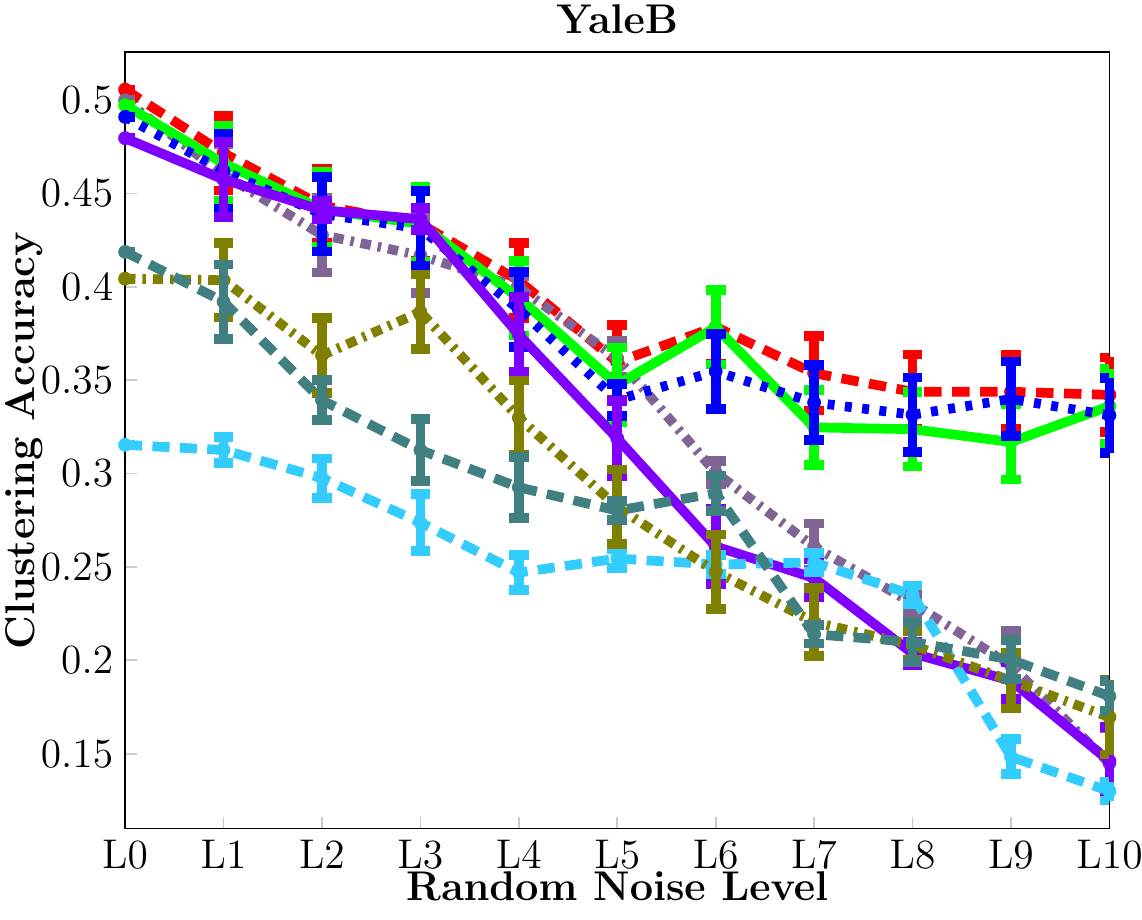} \hspace{-0.26cm}
\includegraphics[width=0.25\linewidth]{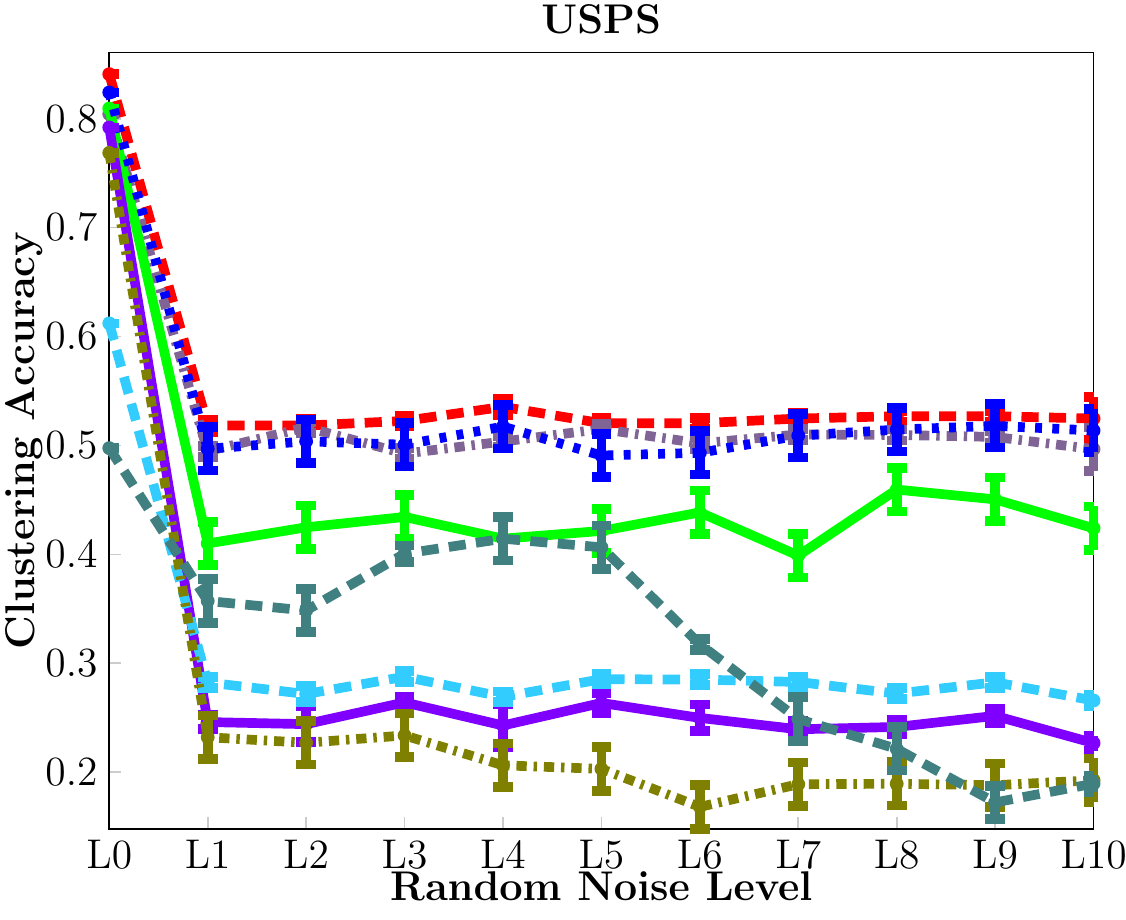} \hspace{-0.26cm}
\includegraphics[width=0.25\linewidth]{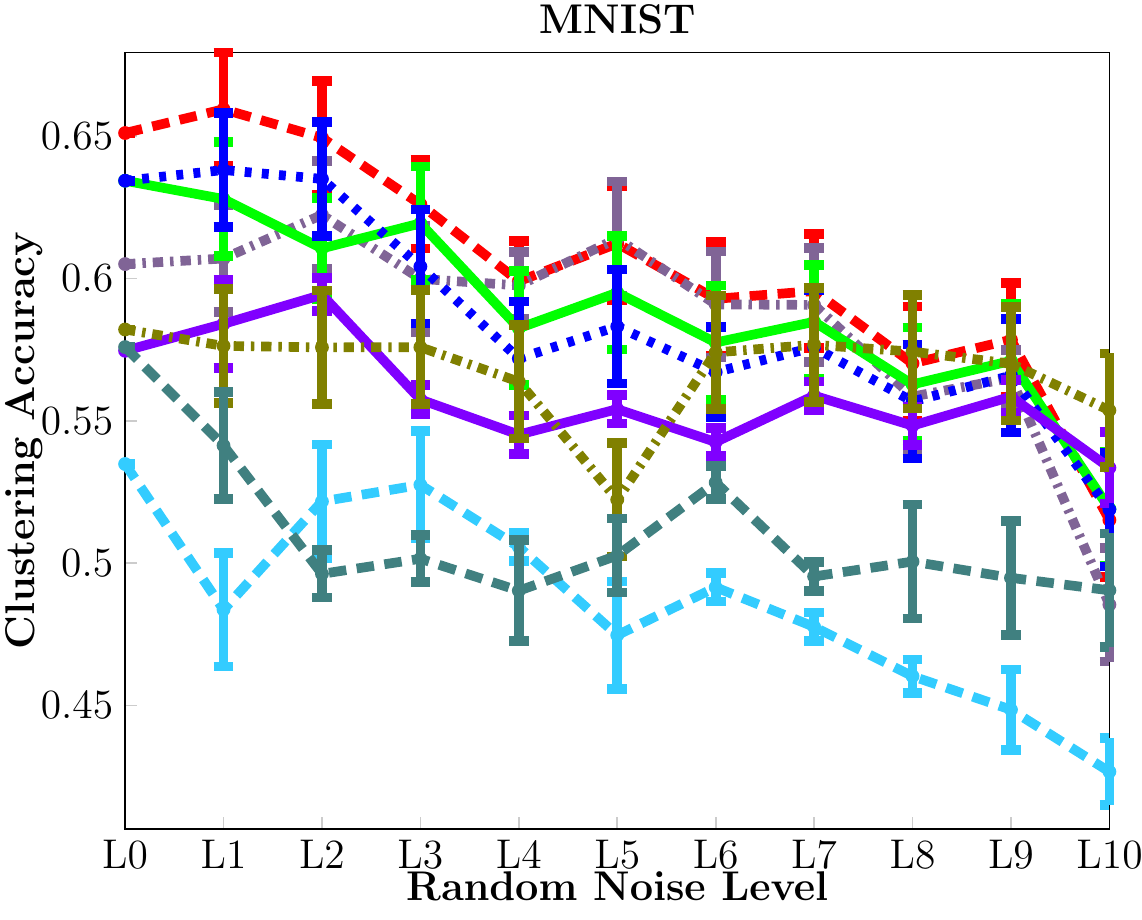}\\
\end{center}
\vspace{-0.28cm}
\caption{Clustering performances of the eight clustering algorithms regarding different random noise levels on the first four datasets.
The first row corresponds to their clustering performances in NMI; and the second row is associated with their clustering performances in accuracy.
The $(\alpha, \beta)$ configurations for weighting different types of hypergraphs on the four datasets are
as follows: (ORL, YaleB.)$\rightarrow$(0.6, 0.2) and (USPS, MNIST)$\rightarrow$(0.4,0.4).
}
\label{fig:face_and_digit_noise_v2} \vspace{-0.31cm}
\end{figure*}

\begin{figure*}[t]
\vspace{-0.15cm}
\begin{center}
\includegraphics[width=0.86\linewidth]{legend.png}\\
\includegraphics[width=0.32\linewidth]{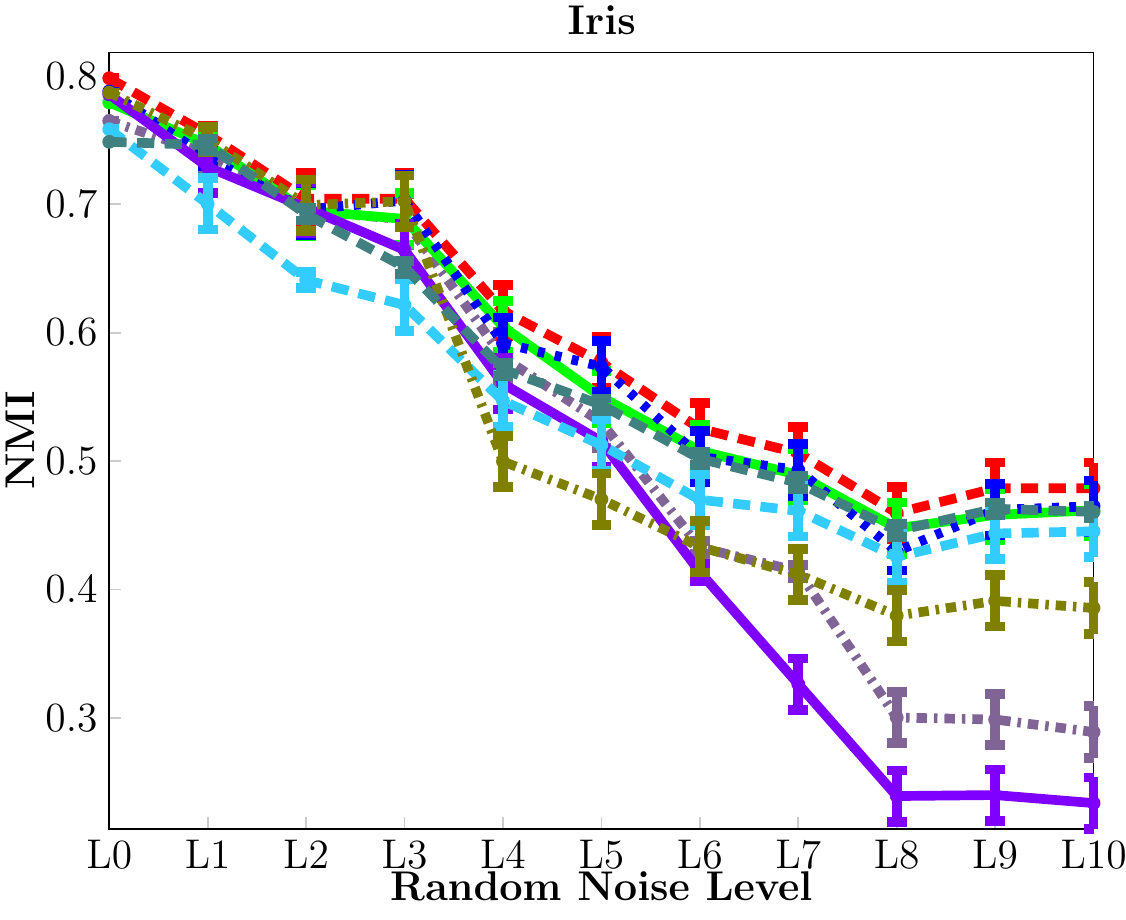} \hspace{-0.26cm}
\includegraphics[width=0.32\linewidth]{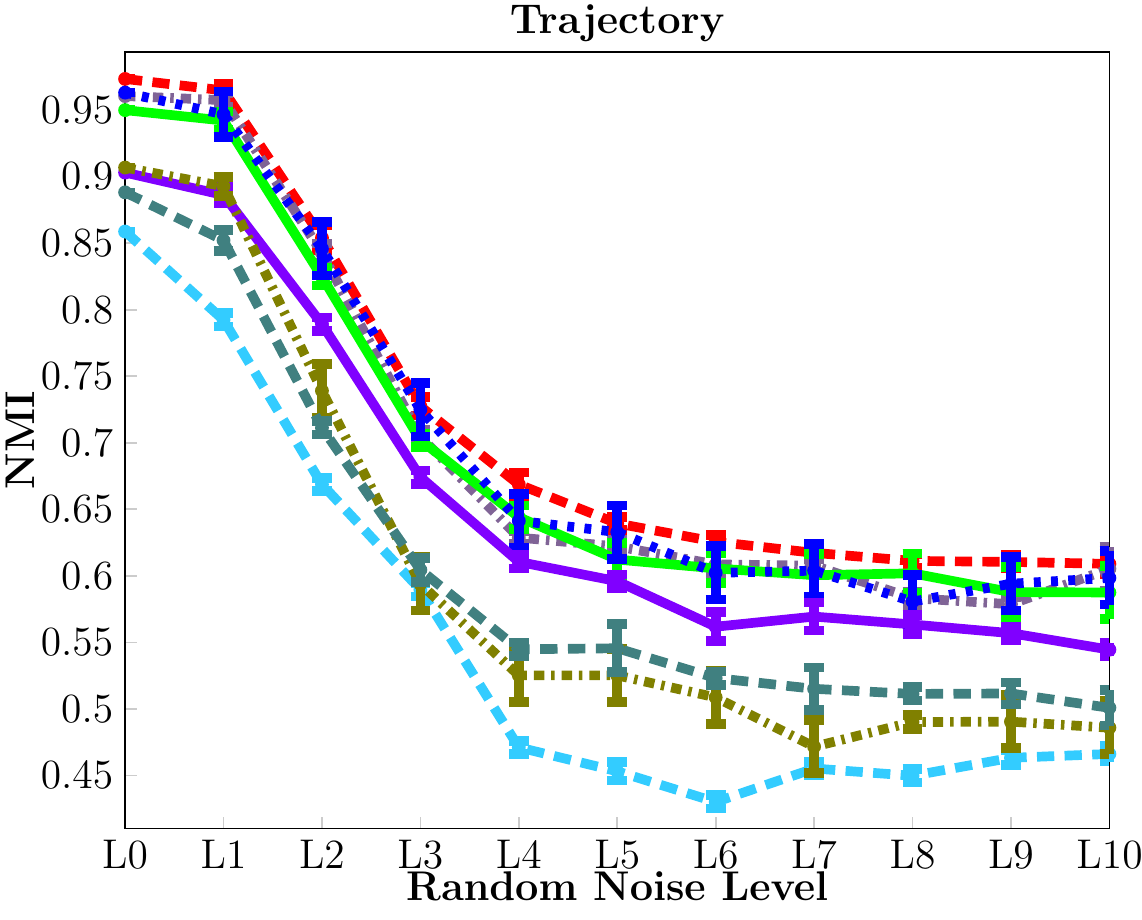} \hspace{-0.26cm}
\includegraphics[width=0.32\linewidth]{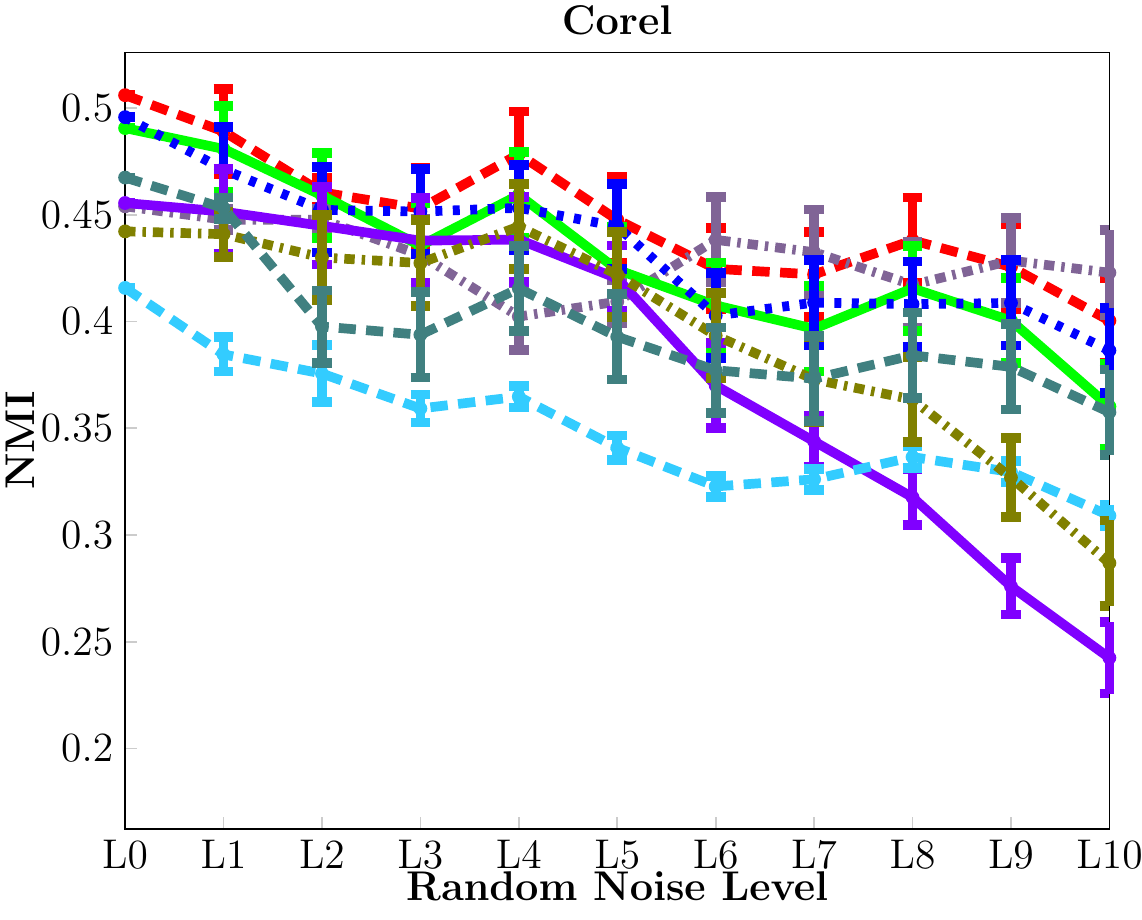}\\ \hspace{-0.26cm}
\includegraphics[width=0.32\linewidth]{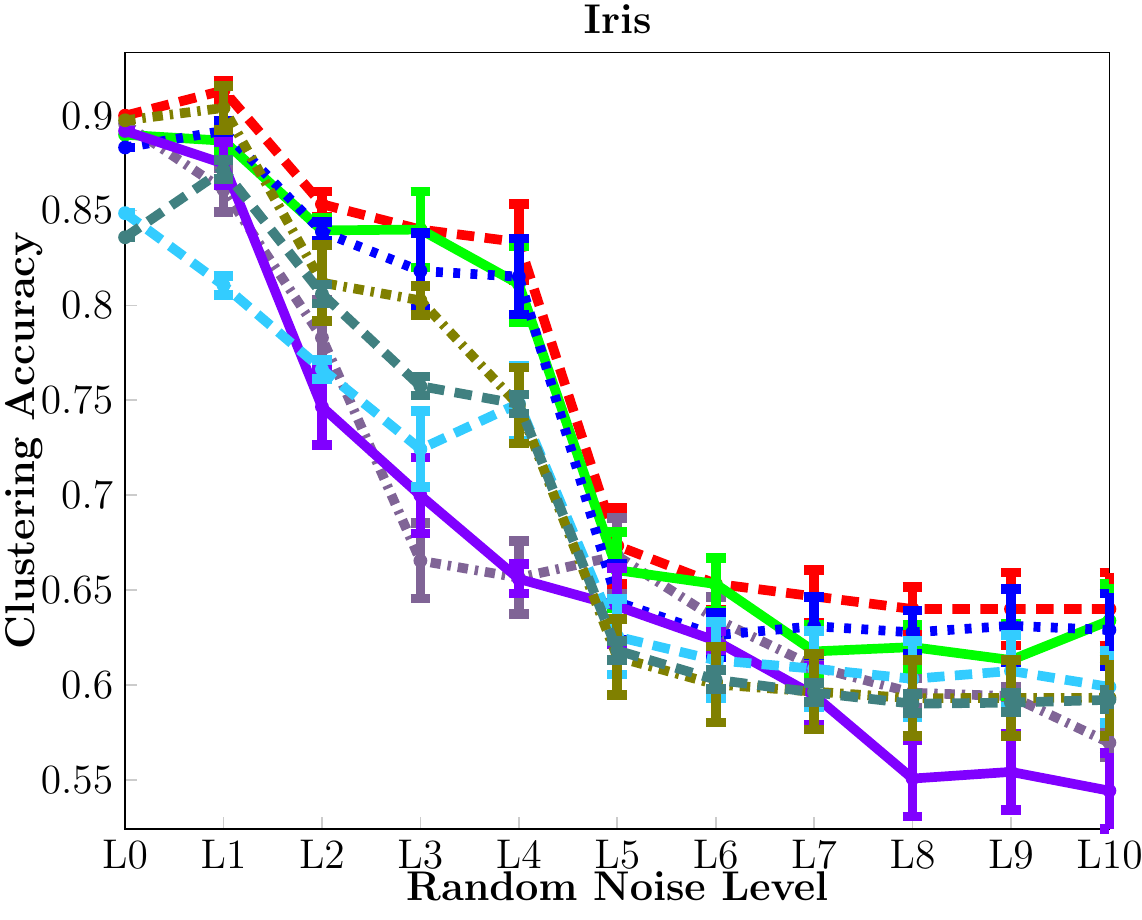} \hspace{-0.26cm}
\includegraphics[width=0.32\linewidth]{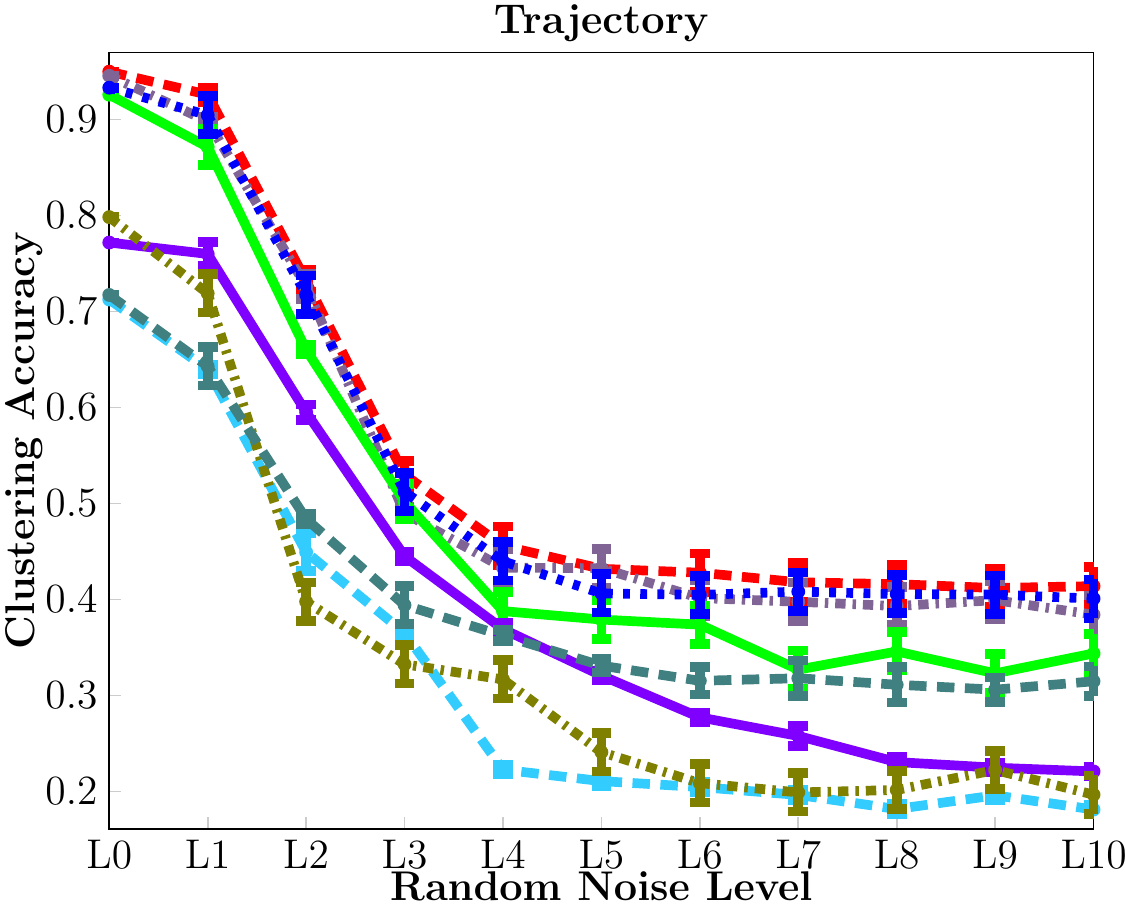}  \hspace{-0.26cm}
\includegraphics[width=0.32\linewidth]{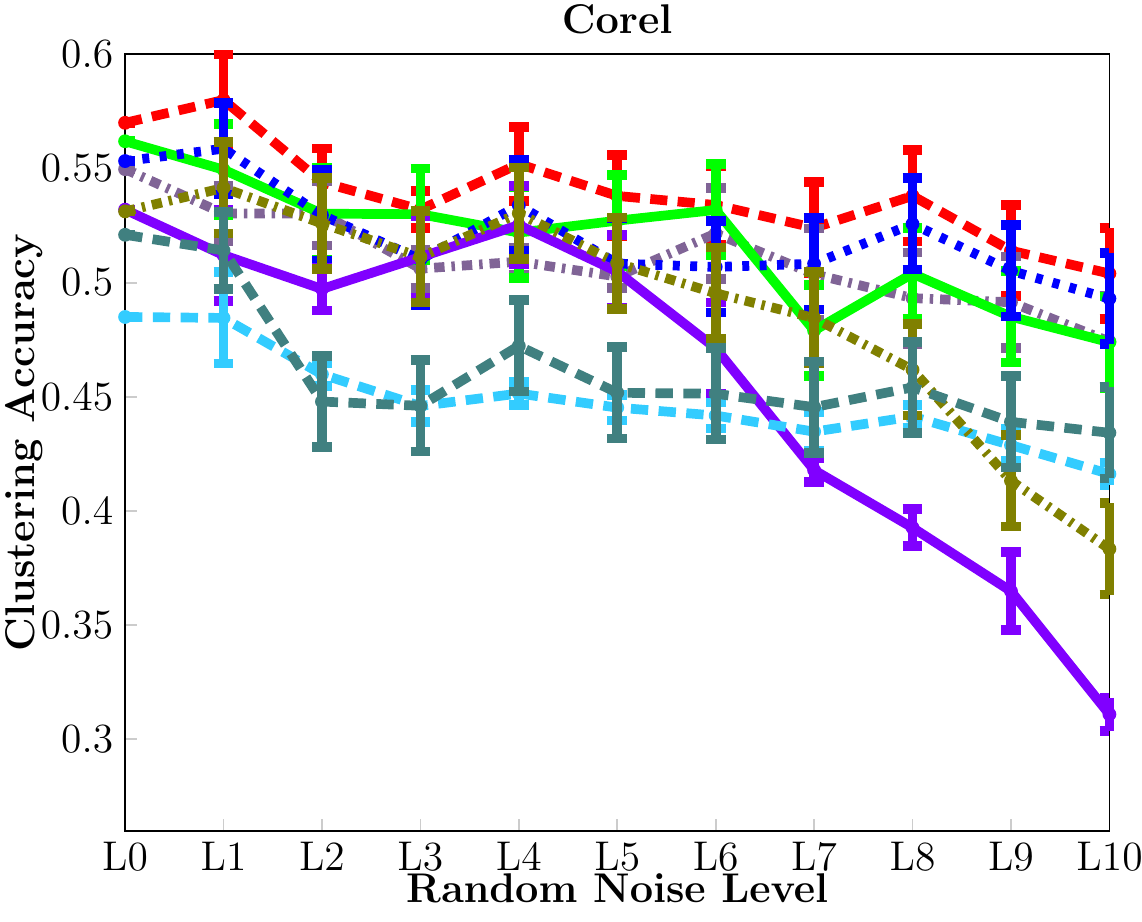}
\end{center}
\vspace{-0.25cm}
\caption{Clustering performances of the eight clustering algorithms regarding different random noise levels on the last three datasets.
The first row corresponds to their clustering performances in NMI; and the second row is associated with their clustering performances in accuracy.
The $(\alpha, \beta)$ configurations for weighting different types of hypergraphs on the three datasets are
as follows: (Traj.)$\rightarrow$(0.6, 0.2) and (Iris, Corel)$\rightarrow$(0.4,0.4).
}
\label{fig:others_noise_v2} \vspace{-0.25cm}
\end{figure*}

\section{Experiments \label{sec:Experiment}}

\subsection{Data description and implementation details}

In the experiments, we evaluate the proposed PKO+HSC on seven datasets, which have the ground truth
labels for classification and clustering tasks. The detailed configurations
of these datasets are given as follows.

\begin{table}
\scriptsize
\scalebox{0.96}{
\begin{tabular}{c|c|c|c|c|c|c|c}
\hline
 & \makebox[0.5cm]{ORL} & \makebox[0.5cm]{YaleB} & \makebox[0.5cm]{USPS} & \makebox[0.5cm]{MNIST} & \makebox[0.5cm]{Iris}  & \makebox[0.5cm]{Traj.} & \makebox[0.5cm]{Corel}\\
\hline
PKO+HSC & \bf 0.8529 & \bf  0.6336 &  \bf 0.7918 & \bf 0.5672 &  \bf 0.7981 &  \bf 0.9734 & \bf 0.5061 \\
PK+HSC & 0.8159 &  0.6306 &  0.7334 &  0.4953 &  0.7648 &  0.9606 &  0.4540 \\
PO+HSC & 0.8111 &  0.6186 &  0.7348 &  0.5583 &  0.7793 &  0.9500 &  0.4907 \\
PKO+NC & 0.8164 &  0.6084 &  0.7815 &  0.5569 &  0.7878 &  0.9631 &  0.4958 \\
CSC & 0.8116 &  0.5849 &  0.7118 &  0.4882 &  0.7855 &  0.9031 &  0.4556 \\
NRSC & 0.7236 &  0.4718 &  0.5807 &  0.4328 &  0.7583 &  0.8587 &  0.4158 \\
RWDSM & 0.7891 &  0.5558 &  0.6741 &  0.4536 &  0.7869 &  0.9068 &  0.4423 \\
STSC& 0.7421 &  0.5295 &  0.4684 &  0.4368 &  0.7486 &  0.8882 &  0.4675 \\
\hline\hline
PKO+HSC & \bf  0.7300 & \bf 0.5055 & \bf 0.8409 & \bf 0.6511 & \bf 0.9000 & \bf 0.9500 &  \bf 0.5700 \\
PK+HSC & 0.7057 &  0.4998 &  0.8043 &  0.6050 &  0.8957 &  0.9457 &  0.5497 \\
PO+HSC & 0.7074 &  0.4972 &  0.8093 &  0.6343 &  0.8899 &  0.9259 &  0.5619 \\
PKO+NC & 0.6883 &  0.4910 &  0.8241 &  0.6343 &  0.8833 &  0.9333 &  0.5533 \\
CSC & 0.6795 &  0.4796 &  0.7920 &  0.5746 &  0.8920 &  0.7720 &  0.5320 \\
NRSC & 0.6015 &  0.3154 &  0.6121 &  0.5348 &  0.8486 &  0.7126 &  0.4850 \\
RWDSM & 0.6669 &  0.4043 &  0.7686 &  0.5821 &  0.8976 &  0.7984 &  0.5312 \\
STSC& 0.6694 &  0.4187 &  0.4976 &  0.5759 &  0.8359 &  0.7173 &  0.5210 \\
\hline
\end{tabular}
}
\caption{Clustering performances of the eight clustering algorithms in NMI and accuracy on the seven datasets without corruption.
Specifically, the upper part corresponds to NMI; and the lower part is associated with accuracy.
The $(\alpha, \beta)$ configurations for weighting different types of hypergraphs on the seven datasets are
as follows: (ORL, YaleB, Traj.)$\rightarrow$(0.6, 0.2) and (USPS, MNIST, Iris, Corel)$\rightarrow$(0.4,0.4).
\label{Tab:quantitative-without-noise}}
\vspace{-0.15cm}
\end{table}

\begin{table}
\scriptsize
\scalebox{0.8}{
\begin{tabular}{c|c|c|c|c|c|c|c|c|c}
\hline
& \makebox[0.5cm]{(0.2, 0.2)} & \makebox[0.5cm]{(0.2, 0.4)} & \makebox[0.5cm]{(0.2, 0.6)} & \makebox[0.5cm]{(0.4, 0.2)} & \makebox[0.5cm]{(0.4, 0.4)}
 & \makebox[0.5cm]{(0.6, 0.2)} & \makebox[0.5cm]{(0.6, 0.4)} & \makebox[0.5cm]{(0.2, 0.0)} & \makebox[0.5cm]{(0.0, 0.2)} \\
\hline
ORL  &   0.8182  &  0.8185   & 0.8229  &  0.8287  &  0.8350  & \bf 0.8529 & 0.8251 & 0.8463 & 0.8359\\
YaleB  &   0.4730  &  0.5153  &  0.5690  &  0.5733  &  0.6189   & \bf 0.6336 & 0.6227 & 0.4792 & 0.4668\\
USPS   &  0.7523   & 0.7648  &  0.7806  &  0.7779   & \bf 0.7918   & 0.7631 & 0.7331 & 0.7059 & 0.7114\\
MNIST   &  0.5500  &  0.5595  &  0.5578   & 0.5561   & \bf 0.5672  &  0.5543& 0.5228& 0.5600 & 0.5574\\
Iris & 0.7961 & 0.7968 & 0.7971 & 0.7978 & \bf 0.7981 & 0.7980 & 0.7777 &  0.7926 & 0.7958\\
Traj. & 0.9548 & 0.9634 & 0.9638 & 0.9641 & 0.9638 & \bf 0.9734 & 0.9631 & 0.9548 & 0.9668\\
Corel & 0.4561 & 0.4772 & 0.4785 & 0.5033 & \bf 0.5061 & 0.4626 & 0.4489 & 0.4561 & 0.4435\\
\hline\hline
ORL  &  0.6775 &   0.6875   & 0.7000    &0.7175   & 0.7075   & \bf 0.7300 & 0.6950 & 0.7225 & 0.7050\\
YaleB &   0.3271   & 0.3727  &  0.4209  &  0.4209   & 0.4450 &   \bf 0.5055 & 0.5044 & 0.3298 & 0.3271\\
USPS  &   0.8065   & 0.8323   & 0.8409  &  0.8301  &  \bf 0.8409  &  0.8194 & 0.8108 & 0.7677 & 0.7892\\
MNIST &   0.6477   & 0.6479  &  0.6411   & 0.6344   & \bf 0.6511   & 0.6500 & 0.6227  & 0.6244 & 0.6477\\
Iris & 0.8903 & 0.8911 & 0.8942 & 0.8953 & \bf 0.9000 & 0.8923 & 0.8900 & 0.8933 & 0.8967\\
Traj. & 0.9401 & 0.9412 & 0.9422 & 0.9431 & 0.9437 & \bf 0.9500 & 0.9400 & 0.9020 & 0.9220\\
Corel & 0.5640 & 0.5600 & 0.5400 & 0.5660 & \bf 0.5700 & 0.5580 & 0.5400 & 0.5340 & 0.5440\\
\hline
\end{tabular}
}
\caption{Clustering performances of the proposed PKO+HSC with different configurations of $(\alpha, \beta)$ in NMI and accuracy on the seven datasets without corruption.
Specifically, the upper part corresponds to NMI; and the lower part is associated with accuracy.
\label{Tab:parameter_tuning} \vspace{-0.19cm}}
\end{table}

The first dataset
is the ORL face
dataset\footnote[1]{http://www.cl.cam.ac.uk/research/dtg/attarchive/facedatabase.html}.
It comprises 400 face images of 40 persons, and each person has 10 images.
The second dataset is a subset of the YaleB face
dataset\footnote[2]{http://vision.ucsd.edu/~leekc/ExtYaleDatabase/ExtYaleB.html},
and contains 2432 near frontal face images from 38 individuals under different illuminations.
For computational convenience, all the face images from the two datasets
are resized to $32\times32$ pixels.
The third dataset is a subset of
the US Postal Service (USPS) handwritten digit dataset\footnote[3]{http://www.csie.ntu.edu.tw/$\sim$cjlin/libsvmtools/datasets/multiclass.html\#usps},
and consists of 9298 $16\times16$ handwritten digit images from ten clusters.
The fourth dataset is a subset of the MNIST handwritten digit dataset\footnote[4]{http://yann.lecun.com/exdb/mnist/}, and constitutes
2000 $28\times28$ digit images from ten clusters.
As shown in Fig.~\ref{fig:traj_sample}, the fifth dataset~\cite{Hsieh-Yu-Chen-TCSVT2006} is a
trajectory dataset containing
2500 trajectories from 50 clusters, and
each cluster
comprises 50 trajectories with complex shapes.
The sixth dataset
is the Iris dataset from the UCI repository\footnote[5]{http://archive.ics.uci.edu/ml/datasets/Iris}, and contains
150 samples from 3 clusters.
The seventh dataset is the Corel image dataset\footnote[6]{Corel Gallery Magic 65000 (1999), www.corel.com} that is composed of
1000 images from ten clusters, as shown in Fig.~\ref{fig:Corel_sample}.

For graph construction, the features used in the two face datasets and
the two digit datasets are directly flattened into grayscale intensity
column vectors. As a result, the feature dimensions for these four datasets
are 1024 (ORL), 1024 (YaleB), 256 (USPS), and 784 (MNIST), respectively.
In addition, the corresponding image features for the Corel dataset
are the 960-dimensional GIST descriptors (as in~\cite{Oliva-Torralba-IJCV2001-gist})
that are widely used in computer vision and pattern recognition. The
corresponding features for the trajectory dataset are 18-dimensional discrete Fourier
transform (DFT) coefficient features (as
in~\cite{Naftel-Khalid-ICVS2006}).
The feature dimension for the Iris dataset is 4, as shown in the UCI repository.
Moreover, the kernel function
$\mathcal{G}(\mathbf{z}_{i}, \mathbf{z}_{j})$
(as in Eq.~\eqref{eq:pairwise_hypergraph_similarity}) is selected as follows:
$\mathcal{G}(\mathbf{z}_{i}, \mathbf{z}_{j}) = \exp\left(-\|\mathbf{z}_{i}-\mathbf{z}_{j}\|^{2}/2\sigma^{2}\right)$
where $\sigma$ is a scaling factor. In practice, $\sigma$ is tuned from the set
$\{y|y=0.2\rho + (\lambda-1)0.2\rho\}$
where $\lambda$ is a positive integer such that $\lambda \in \{1, 2, \ldots, 15\}$ and
$\rho$ is the average of the distances from each sample
$\mathbf{z}_{i}$ to the other samples. The parameter $k$ in the \KNN hypergraph is set
to 3.
The weighting factors $(\alpha, \beta)$ in Eq.~\eqref{eq:hyper_similarity_measure} are
chosen from the set $\{$(0.2, 0.2),  (0.2, 0.4), (0.2, 0.6), (0.4, 0.2), (0.4, 0.4),
(0.6, 0.2), (0.6, 0.4), (0.2, 0.0), (0.0, 0.2)$\}$. The task of constructing the high-order
over-clustering hypergraph can be accomplished
by using a set of existing clustering methods.
In our case, we
take advantage of
classic spectral clustering~\cite{Ng-Jordan-Weiss15} and
multi-class spectral clustering~\cite{Yu-Shi14}
to generate a set of vertex communities.
For each over-clustering method,
the number of the vertex
communities is chosen as $2\mathcal{K}$ with
$\mathcal{K}$ being the desired number of clusters (referred to in Algorithm~\ref{alg:Framwork}).
Since we focus on the issues of iii) and iv) referred to in Sec.~\ref{sec:intro},
$\mathcal{K}$ is directly set as the ground truth number of clusters for each dataset.
The above experimental
configurations remain the same for all the experiments below.

\vspace{-0.16cm}
\textbf{Computational complexity analysis}
Given $N$ data samples, our pairwise hypergraph construction requires $N^{2}$ kernel computation operations (referred to
in Eq.~\eqref{eq:pairwise_weight}). Accordingly, the $k$NN hypergraph construction needs to calculate $N^{2}$ cosine similarities (defined
in Eq.~\eqref{eq:KNN_hypergraph_similarity_matrix}) with respect to the $N$ data samples. Similarly, the over-clustering hypergraph
construction involves $N^{2}$ cosine similarity computation operations (mentioned in Eq.~\eqref{eq:over-clustering_hypergraph_similarity_matrix}).
The main computational cost of graph partitioning lies in the eigenvalue decomposition of $S - \rho Q$ while solving the optimization problem~\eqref{eq:GraphOptimization5} using the
Newton-Lanczos algorithm. According to~\cite{Lanczos_wiki}, the time complexity of the Lanczos iterations in our graph partitioning is $\mathcal{O}(N^{2})$.
Therefore, the overall time complexity of our method is $\mathcal{O}(N^{2})$, which is the same as standard spectral clustering methods.
For example,
the average running time of our method on the USPS dataset (s.t. $N=9298$) is
15.21 seconds.
The spatial complexity of our method lies in the four aspects: 1)
the $N\times N$ pairwise hypergraph incidence matrix $H_{p}$;
2) the $N\times N$ $k$NN hypergraph incidence matrix $H_{n}$;
3) the $N\times N$ over-clustering hypergraph incidence matrix $H_{o}$;
and 4) the final $N\times N$ similarity matrix $S$. It therefore has $\mathcal{O}(N^{2})$ overall spatial complexity.

\vspace{-0.15cm}
\subsection{Competing algorithms}

We compare the proposed PKO+HSC with
several representative spectral clustering algorithms.
These spectral clustering algorithms are recently proposed, and have significant impacts
on the data clustering community. For descriptive convenience, they are respectively referred to as
CSC (classic spectral clustering~\cite{Ng-Jordan-Weiss15}),
STSC (self-tuning spectral clustering~\cite{Zelnik-Manor-Perona-NIPS2005}),  and NRSC (noise-robust spectral clustering~\cite{Li-Liu-Chen-Tang-ICCV2007}).

In order to verify the effect of different hypergraph components, we compare
the proposed PKO+HSC with
PK+HSC (pairwise+\KNN hypergraph spectral clustering)
and PO+HSC (pairwise+over-clustering hypergraph spectral clustering).
Actually,  PK+HSC and PO+HSC are special cases of the proposed PKO+HSC with different configurations
of $(\alpha, \beta)$.
In order to evaluate the performance of different graph partitioning
criteria, we perform a comparison experiment against
PKO+NC (our context-aware hypergraph
similarity measure together with the normalized
cut criterion~\cite{Yu-Shi14}).
Furthermore, in order to demonstrate the effectiveness of our context-aware hypergraph
similarity measure (defined in Eq.~\eqref{eq:hyper_similarity_measure}), we
make a quantitative comparison with another similarity measure called
RWDSM (Random Walk Diffusion Similarity Measure~\cite{Li-ACCV2009-SC}).
We put both RWDSM and our similarity measure into the same
discriminative hypergraph partitioning criterion (DHPC)
for data clustering.

\begin{table}[t]
\scriptsize
\scalebox{0.96}{
\begin{tabular}{c|c|c|c|c|c|c|c}
\hline
 & \makebox[0.25cm]{ORL} & \makebox[0.25cm]{YaleB} & \makebox[0.25cm]{USPS} & \makebox[0.25cm]{MNIST} & \makebox[0.25cm]{Iris}  & \makebox[0.25cm]{Traj.} & \makebox[0.25cm]{Corel}\\
\hline
PKO+HSC& \bf 0.7626 & \bf 0.5328 & \bf 0.4966 & \bf 0.5148 & \bf 0.6003 & \bf 0.7183 & \bf 0.4496 \\
PK+HSC& 0.7266 &  0.4903 &  0.4834 &  0.4916 &  0.5212 &  0.7000 &  0.4302 \\
PO+HSC& 0.7370 &  0.5191 &  0.4052 &  0.5019 &  0.5844 &  0.6962 &  0.4300 \\
PKO+NC& 0.7453 &  0.5144 &  0.4819 &  0.4988 &  0.5856 &  0.7031 &  0.4349 \\
CSC& 0.6935 &  0.4634 &  0.1502 &  0.4589 &  0.4914 &  0.6596 &  0.3817 \\
NRSC& 0.6379 &  0.3853 &  0.1723 &  0.3798 &  0.5478 &  0.5544 &  0.3514 \\
RWDSM& 0.6493 &  0.4441 &  0.1308 &  0.4508 &  0.5373 &  0.6027 &  0.3955 \\
STSC& 0.6494 &  0.4315 &  0.3260 &  0.3944 &  0.5734 &  0.6099 &  0.3992 \\
\hline\hline
PKO+HSC& \bf 0.6107 & \bf 0.3981 & \bf 0.5527 & \bf 0.6045 & \bf 0.7485 & \bf 0.5558 & \bf 0.5391 \\
PK+HSC& 0.5614 &  0.3364 &  0.5321 &  0.5851 &  0.6849 &  0.5366 &  0.5102 \\
PO+HSC& 0.5725 &  0.3872 &  0.4624 &  0.5895 &  0.7333 &  0.4946 &  0.5177 \\
PKO+NC& 0.5884 &  0.3860 &  0.5348 &  0.5865 &  0.7306 &  0.5399 &  0.5212 \\
CSC& 0.4988 &  0.3229 &  0.2966 &  0.5592 &  0.6707 &  0.4067 &  0.4582 \\
NRSC& 0.4650 &  0.2471 &  0.3089 &  0.4866 &  0.6868 &  0.3237 &  0.4486 \\
RWDSM& 0.4558 &  0.2913 &  0.2545 &  0.5677 &  0.7049 &  0.3486 &  0.4897 \\
STSC& 0.5429 &  0.2845 &  0.3250 &  0.5107 &  0.6917 &  0.4089 &  0.4616 \\
\hline
\end{tabular}
}
\caption{Clustering performances of the eight clustering algorithms in NMI and accuracy on the seven datasets by averaging different random noise levels.
Specifically, the upper part corresponds to NMI; and the lower part is associated with accuracy.
The $(\alpha, \beta)$ configurations for weighting different types of hypergraphs on the seven datasets are
as follows: (ORL, YaleB, Traj.)$\rightarrow$(0.6, 0.2) and (USPS, MNIST, Iris, Corel)$\rightarrow$(0.4,0.4).
\label{Tab:quantitative_with_noise} \vspace{-0.38cm}}
\end{table}

\vspace{-0.15cm}
\subsection{Evaluation criteria}

For a quantitative comparison, we introduce two evaluation criteria---NMI (normalized mutual information)
and clustering accuracy. Specifically, the NMI criterion is defined as:
${\rm NMI}(\mathbf{X}, \mathbf{Y}) = I(\mathbf{X},
\mathbf{Y})/\sqrt{H(\mathbf{X})H(\mathbf{Y})}$
where $\mathbf{X}$ and $\mathbf{Y}$ are two random variables, $H(\mathbf{X})$ and $H(\mathbf{Y})$
are the corresponding entropies of $\mathbf{X}$ and $\mathbf{Y}$, and
$I(\mathbf{X}, \mathbf{Y})$
is the mutual information on $\mathbf{X}$ and $\mathbf{Y}$. In
principle, $\mbox{NMI}(\mathbf{X}, \mathbf{Y})$
has the range of $[0, 1]$, and is equal to 1 when $\mathbf{X}=\mathbf{Y}$.
As far as data clustering is concerned, the NMI criterion is explicitly formulated as: \vspace{-0.3cm}
\begin{equation}
\mbox{NMI}(\mathbf{S}', \mathbf{S})
=\frac{\sum_{i=1}^{\mathcal{C}}\sum_{j=1}^{K}\frac{q_{ij}}{m}\log \left(\frac{\frac{q_{ij}}{m}}{\frac{m_{i}}{m}\frac{m_{j}'}{m}}\right)}
{\sqrt{(\sum_{i=1}^{\mathcal{C}}\frac{m_{i}}{m}\log \frac{m_{i}}{m})(\sum_{j=1}^{K}\frac{m'_{j}}{m}\log \frac{m'_{j}}{m})}} \vspace{-0.15cm}
\end{equation}
where $\mathbf{S}=\{S_{i}\}_{i=1}^{\mathcal{C}}$ is the ground truth clustering configuration of a dataset,
$\mathbf{S}'=\{S'_{j}\}_{j=1}^{K}$ is the clustering configuration obtained by a clustering algorithm,
$\mathcal{C}$ is the ground truth cluster number, $K$ is the obtained cluster number,
$q_{ij}$ is the cardinality of the intersection of $S_{i}$ and $S'_{j}$,
$m_{i}$ is the cardinality of $S_{i}$, $m'_{j}$ is the cardinality of $S'_{j}$,
and $m$ is the cardinality of the whole dataset.  The larger the NMI, the better the clustering performance.

On the other hand, the clustering accuracy is defined as:
$\mbox{Accuracy} = \frac{1}{m}\sum_{j=1}^{K}n_{j}$
where $n_{j}$ is the number of
the samples whose ground truth cluster labels have the highest proportion
in the $j$-th cluster $S'_{j}$. The larger the clustering accuracy, the better
the clustering results.

\vspace{-0.15cm}
\subsection{Clustering results}

In the experiments, we aim to evaluate the clustering performances of different
clustering algorithms in the following three aspects:
i) evaluating the clustering performances (i.e., NMI and clustering accuracy) on the original datasets;
ii) quantitative comparisons on the datasets after noise-based feature perturbations;
and iii) performance evaluations on the datasets with outlier corruptions.
The purposes of the above-mentioned three aspects are to verify the clustering effectiveness
and the clustering robustness.

\begin{figure*}[t]
\vspace{-0.0cm}
\begin{center}
\includegraphics[width=0.85\linewidth]{legend.png}\\
\includegraphics[width=0.236\linewidth]{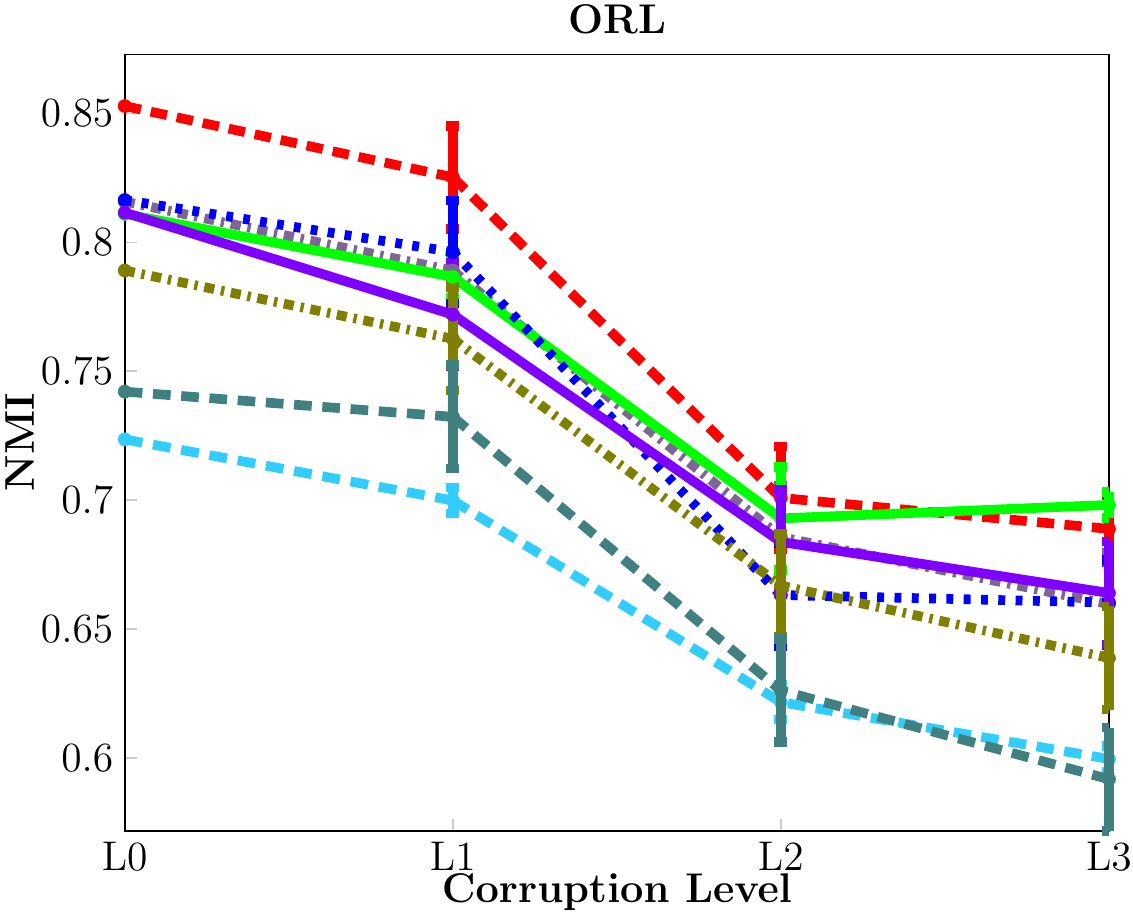} \hspace{-0.26cm}
\includegraphics[width=0.236\linewidth]{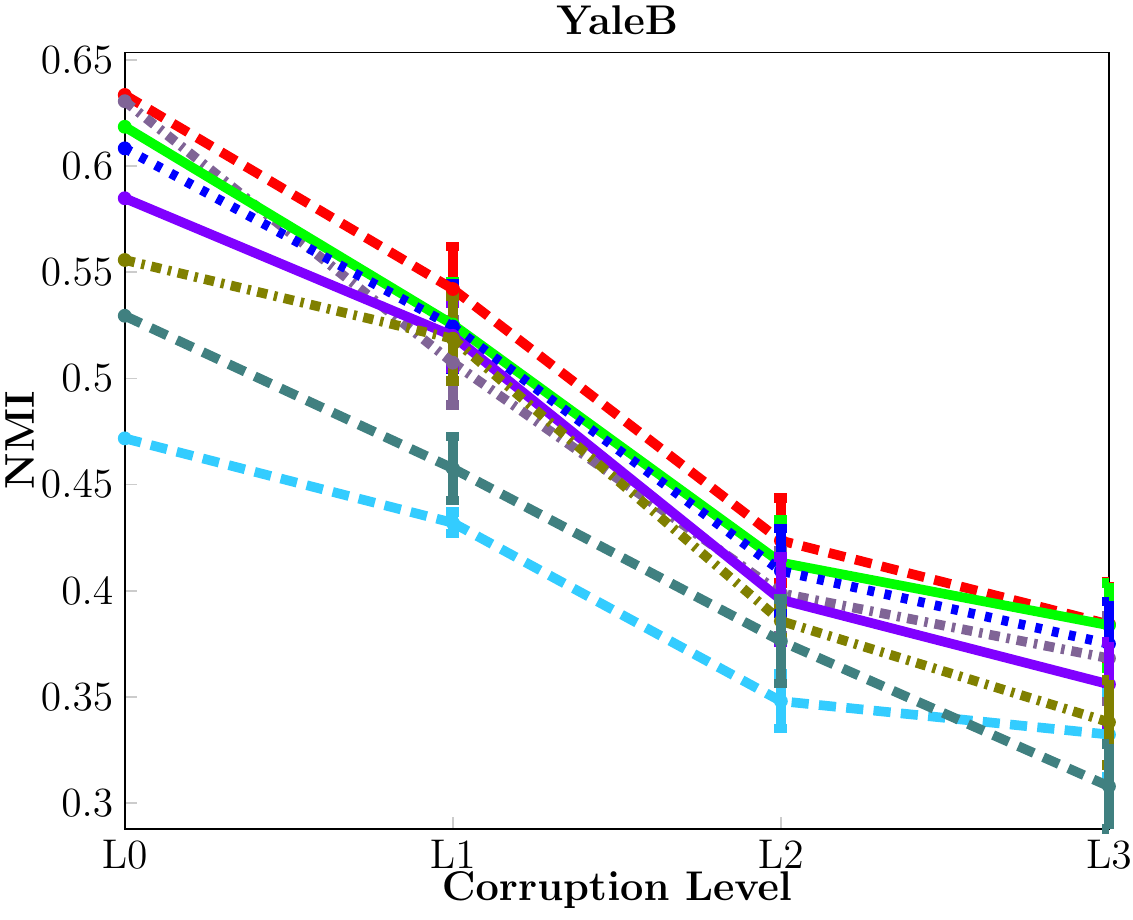} \hspace{-0.26cm}
\includegraphics[width=0.236\linewidth]{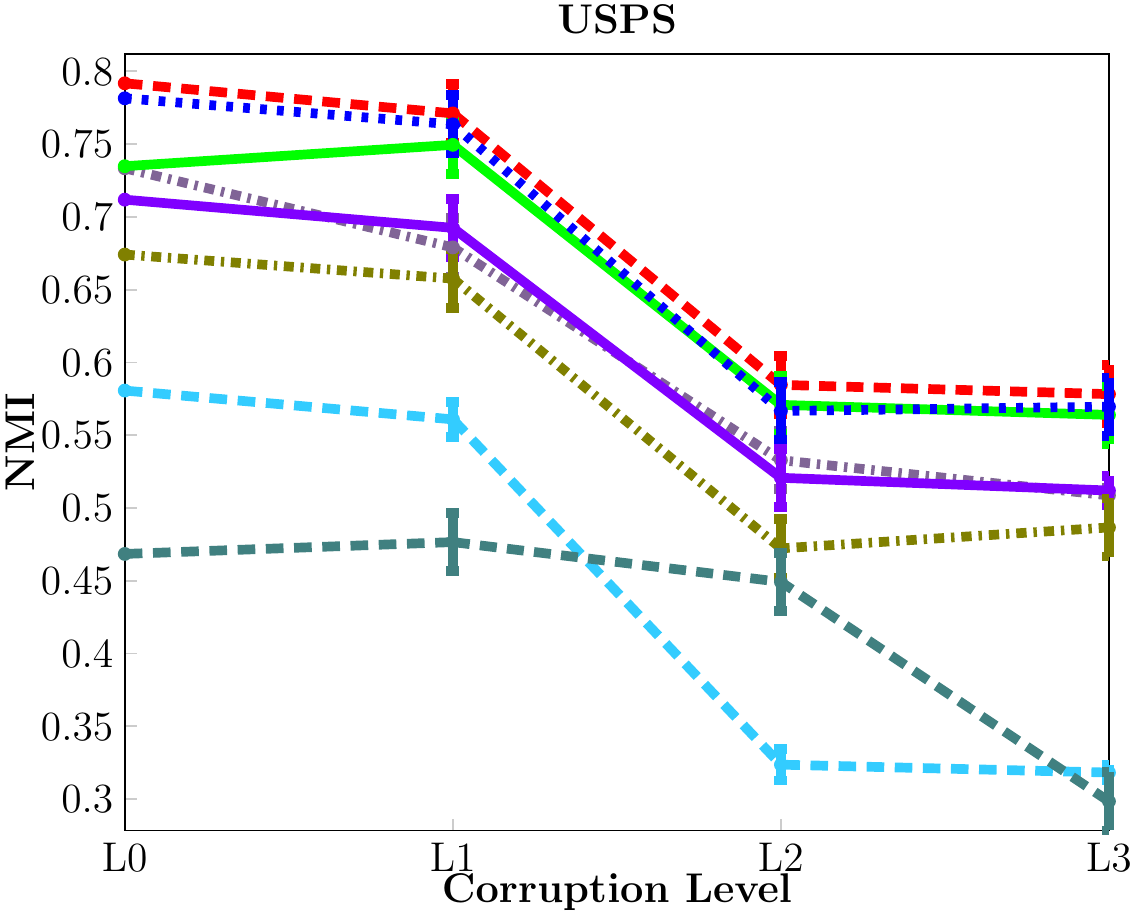} \hspace{-0.26cm}
\includegraphics[width=0.236\linewidth]{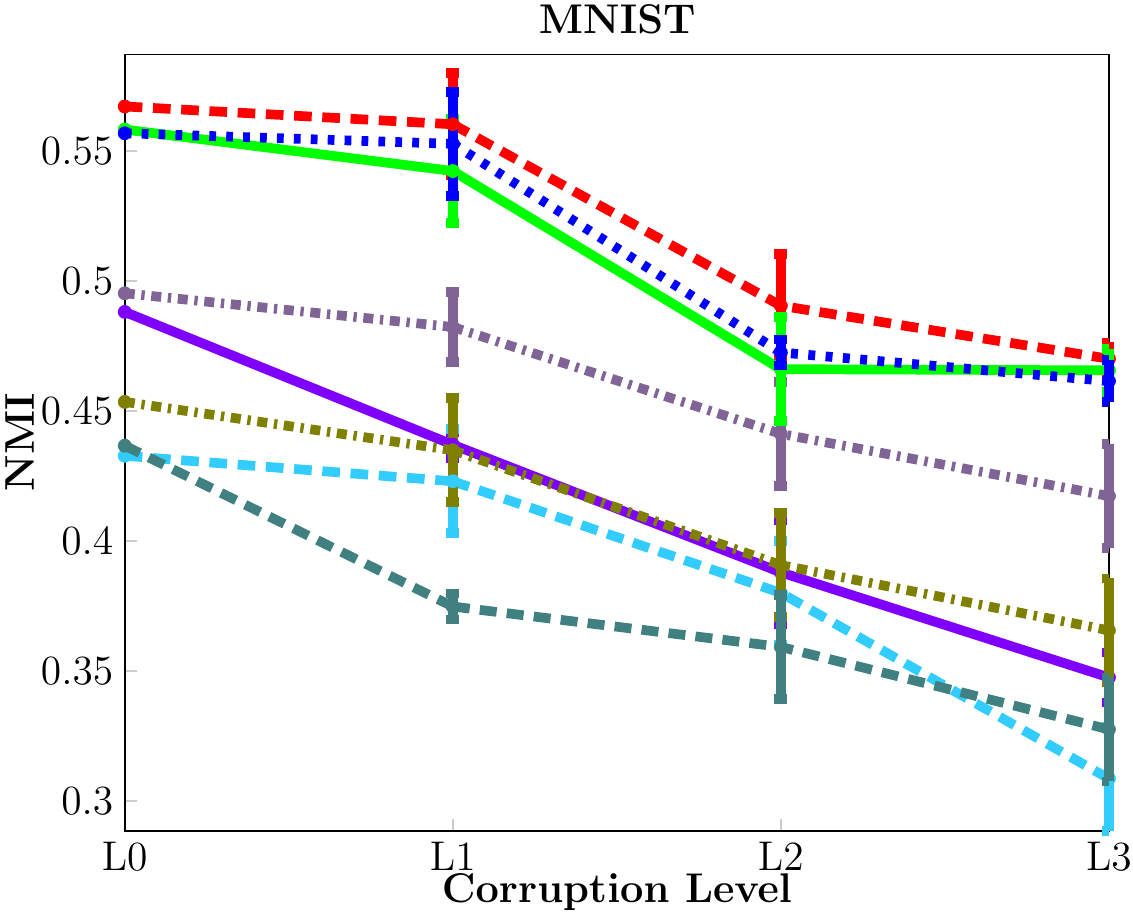}\\
\includegraphics[width=0.236\linewidth]{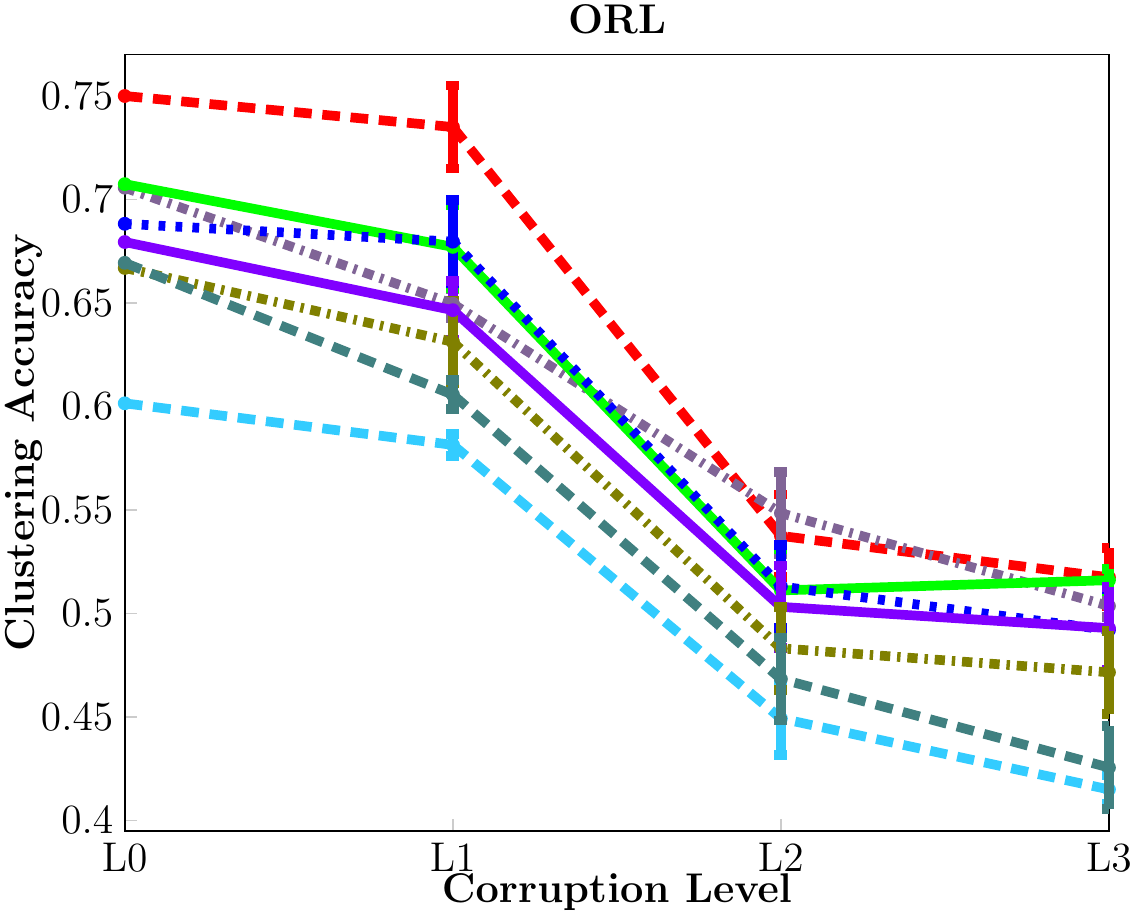} \hspace{-0.26cm}
\includegraphics[width=0.236\linewidth]{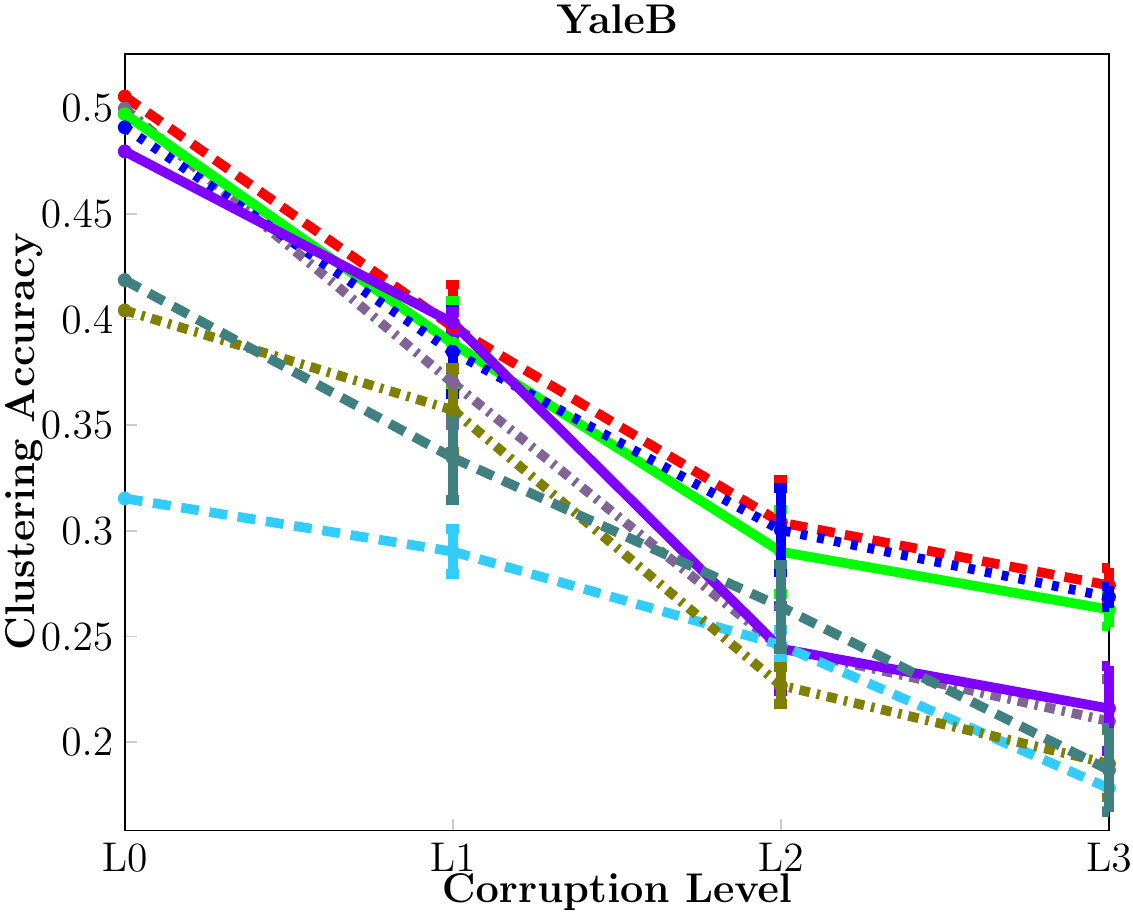} \hspace{-0.26cm}
\includegraphics[width=0.236\linewidth]{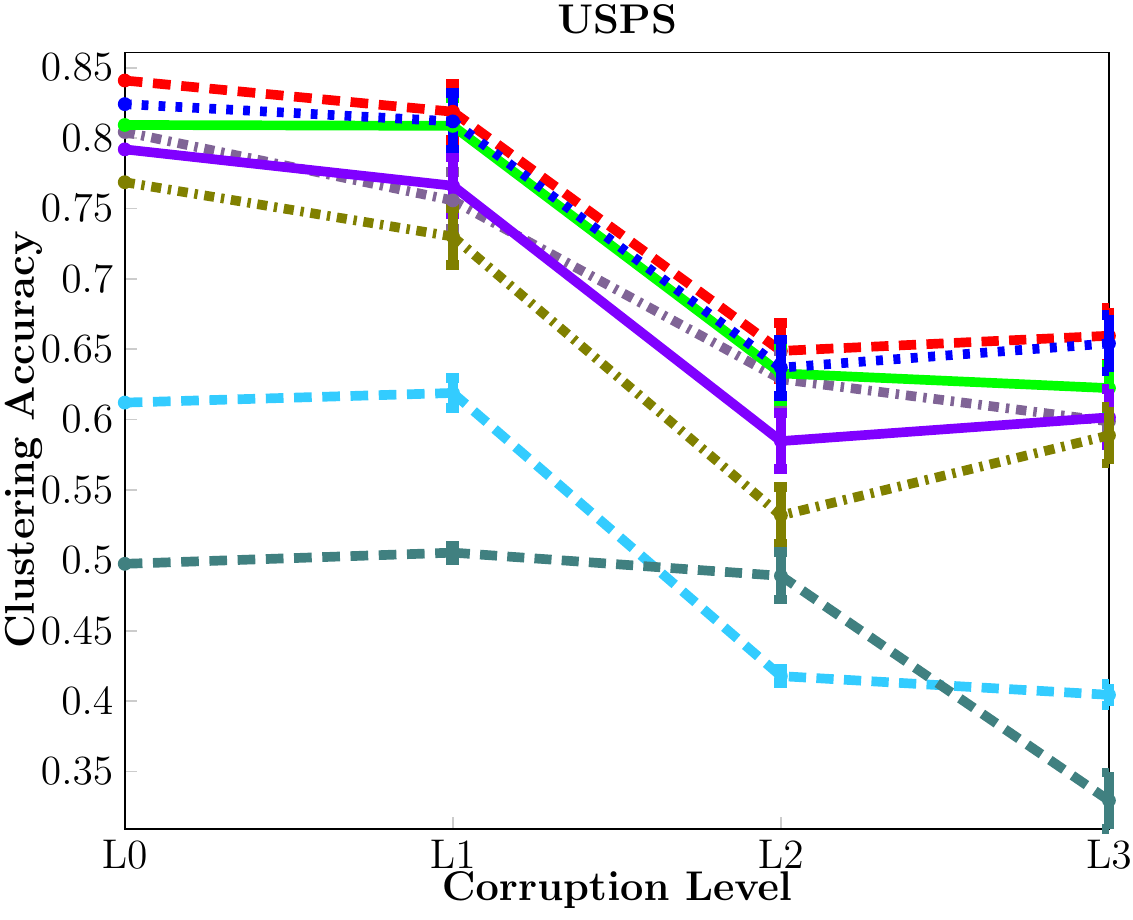} \hspace{-0.26cm}
\includegraphics[width=0.236\linewidth]{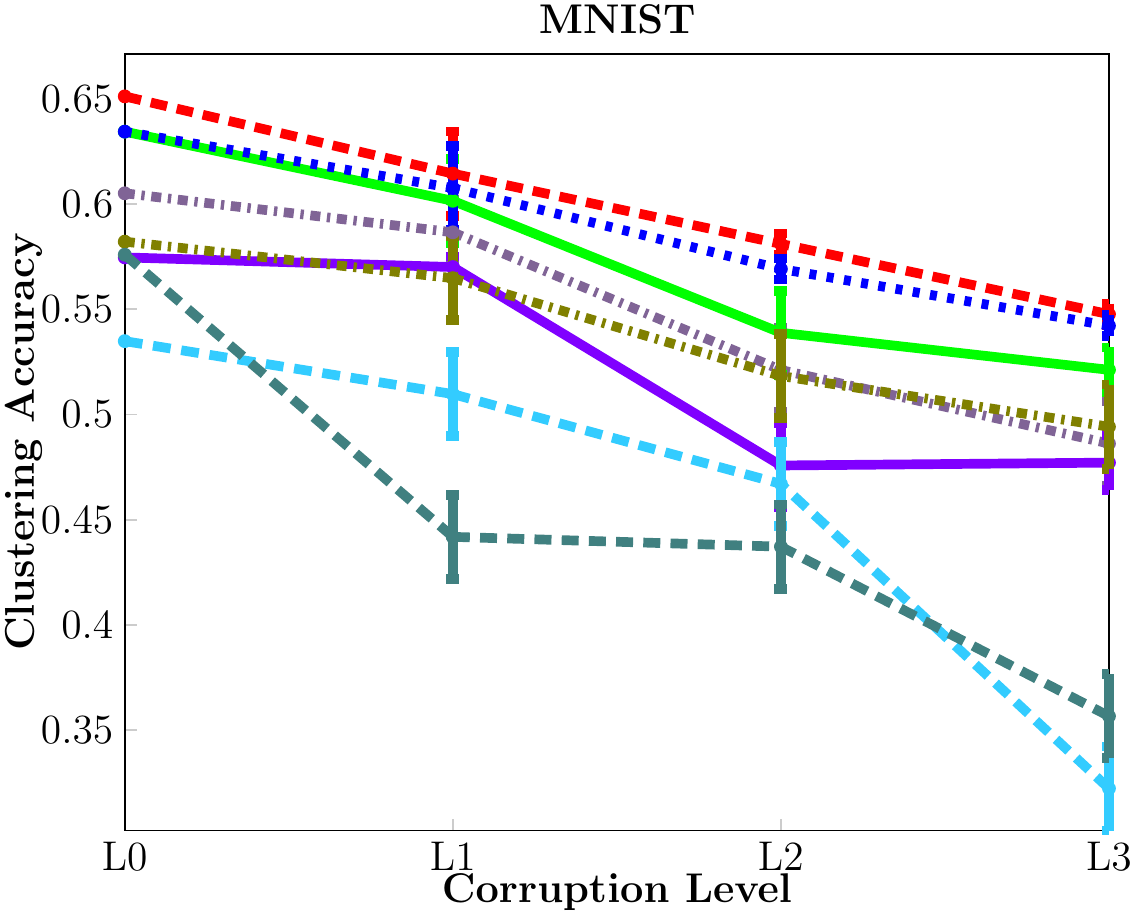}\\
\end{center}
\vspace{-0.35cm}
\caption{Clustering performances of the eight clustering algorithms regarding different outlier corruption levels on the first four datasets.
The first row corresponds to their clustering performances in NMI; and the second row is associated with their clustering performances in accuracy.
The $(\alpha, \beta)$ configurations for weighting different types of hypergraphs on the four datasets are
as follows: (ORL, YaleB.)$\rightarrow$(0.6, 0.2) and (USPS, MNIST)$\rightarrow$(0.4,0.4).
}
\label{fig:face_and_digit_noise_v3} \vspace{-0.36cm}
\end{figure*}

For i), Tab.~\ref{Tab:quantitative-without-noise} reports the corresponding NMIs and accuracies
of the eight clustering algorithms. It is seen from Tab.~\ref{Tab:quantitative-without-noise}
that the proposed PKO+HSC obtains higher NMIs and accuracies than the other clustering
algorithms.  More specifically, the average NMI gains of PKO+HSC
regarding the seven datasets are
(5.53\%, 3.65\%, 2.26\%, 8.07\%, 20.78\%, 11.16\%, 19.67\%)
over
those of (PK+HSC, PO+HSC, PKO+NC, CSC, NRSC, RWDSM, STSC), respectively;
and the average accuracy gains are
(2.83\%, 2.42\%, 2.80\%, 9.02\%, 25.55\%, 10.72\%, 21.58\%),
respectively.
Furthermore, Tab.~\ref{Tab:parameter_tuning} reports the NMIs and accuracies
of the proposed PKO+HSC with different configurations of the weighting factors $(\alpha, \beta)$.
From Tab.~\ref{Tab:parameter_tuning}, we see that the clustering performances of the proposed PKO+HSC are
not very sensitive to the configurations of the weighting factors.

For ii), noise-based feature perturbations are performed by using additive random noises.
Figs.~\ref{fig:face_and_digit_noise_v2} and~\ref{fig:others_noise_v2} show
the NMI and accuracy curves with error bars in eleven different noise levels
(i.e., $\mbox{L}_{0}$ corresponds to i), and $\mbox{L}_{1}\rightarrow \mbox{L}_{10}$ are
associated with ten ascending noise levels whose magnitudes
are chosen from $\{0.2, 0.4, 0.6, 0.8, 1.0, 1.2, 1.4, 1.6, 1.8, 2.0\}$).
Clearly,  the proposed PKO+HSC
achieves the highest NMIs and accuracies in most noise levels.
Furthermore, Tab.~\ref{Tab:quantitative_with_noise} reports the average
NMIs and accuracies of the eight clustering algorithms regarding different
noise levels. The average NMI and accuracy gains of the proposed PKO+HSC
are
(6.03\%, 5.19\%, 2.80\%, 23.54\%, 34.54\%, 26.92\%, 20.43\%)
and
(7.01\%, 6.71\%, 3.14\%, 24.78\%, 35.19\%, 28.81\%, 24.32\%)
over those of (PK+HSC, PO+HSC, PKO+NC, CSC, NRSC, RWDSM, STSC), respectively.

For iii), outlier corruptions are performed by randomly setting the feature elements to zeros.
Fig.~\ref{fig:face_and_digit_noise_v3} displays
the clustering performances of the eight clustering algorithms regarding four outlier corruption levels
(i.e., $\mbox{L}_{0}$ corresponds to i), and $\mbox{L}_{1}\rightarrow \mbox{L}_{3}$ are
associated with three ascending corruption levels
whose corruption ratios are chosen from $\{0.2, 0.4, 0.6\}$).
From Fig.~\ref{fig:face_and_digit_noise_v3}, we see that the proposed PKO+HSC consistently achieves higher NMIs and accuracies than
the other clustering algorithms.  Moreover, Tab.~\ref{Tab:face_and_digit_noise_v3} shows the
average
NMIs and accuracies of the eight clustering algorithms regarding different
outlier corruption levels. The average NMI and accuracy gains of the proposed PKO+HSC
are
(7.87\%, 2.96\%, 2.99\%, 11.02\%, 30.55\%, 14.83\%, 30.58\%)
and
(7.39\%, 3.80\%, 2.91\%, 10.11\%, 34.62\%, 14.04\%, 33.66\%)
over those of (PK+HSC, PO+HSC, PKO+NC, CSC, NRSC, RWDSM, STSC), respectively.

\begin{figure}[t]
\vspace{-0.1cm}
\begin{center}
\includegraphics[width=0.98\linewidth]{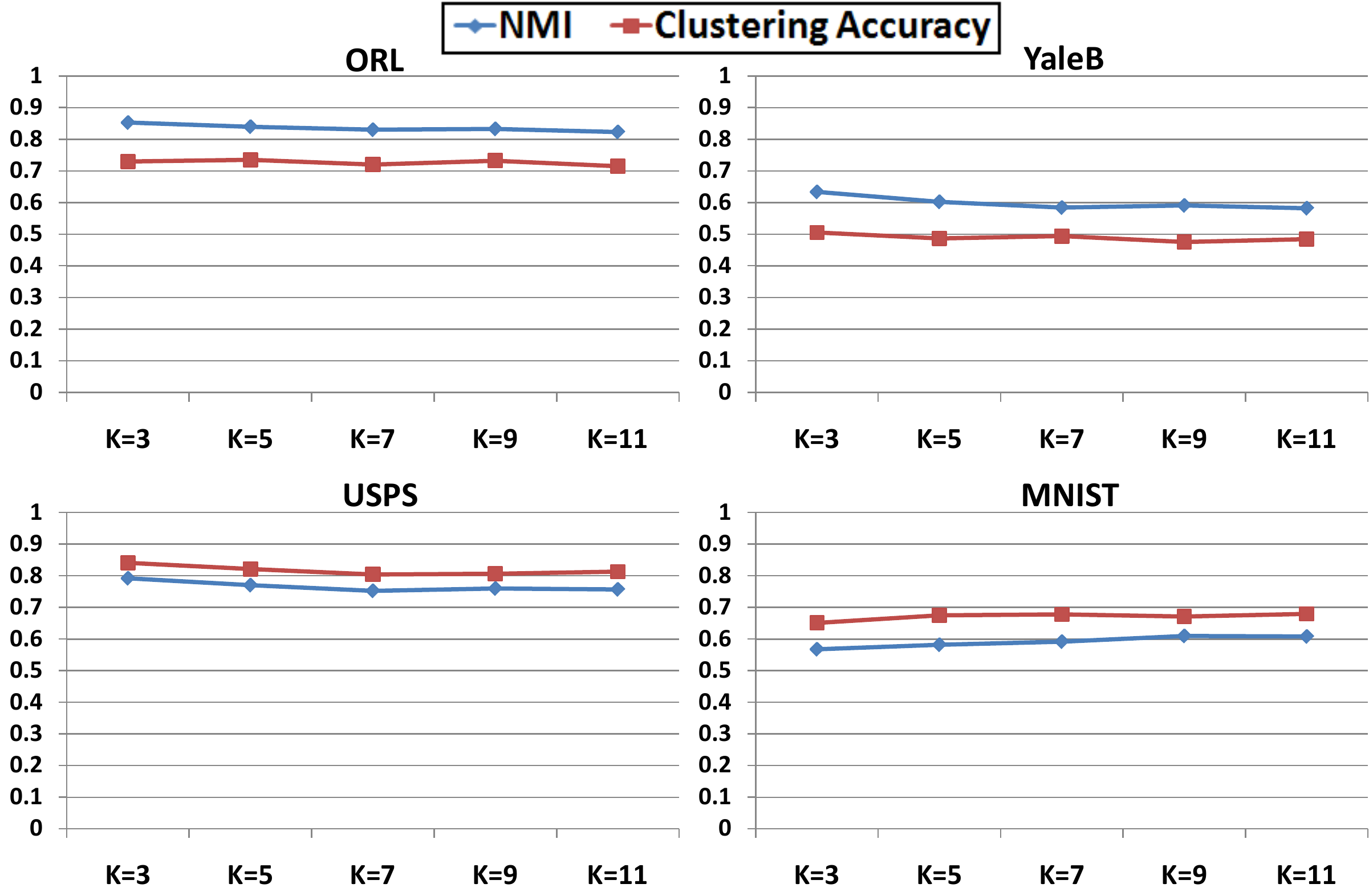}
\end{center}
\vspace{-0.19cm}
\caption{Clustering performances of the proposed PKO+HSC with different configurations
of the nearest-neighbor number $k$ for \KNN hypergraph construction
in NMI and accuracy on the four datasets without corruption.
The $(\alpha, \beta)$ configurations for weighting different types of hypergraphs on the four datasets are
as follows: (ORL, YaleB.)$\rightarrow$(0.6, 0.2) and (USPS, MNIST)$\rightarrow$(0.4,0.4).
}
\label{fig:ORL_K_Selection} \vspace{-0.6cm}
\end{figure}

\begin{figure}[t]
\vspace{-0.1cm}
\begin{center}
\includegraphics[width=0.98\linewidth]{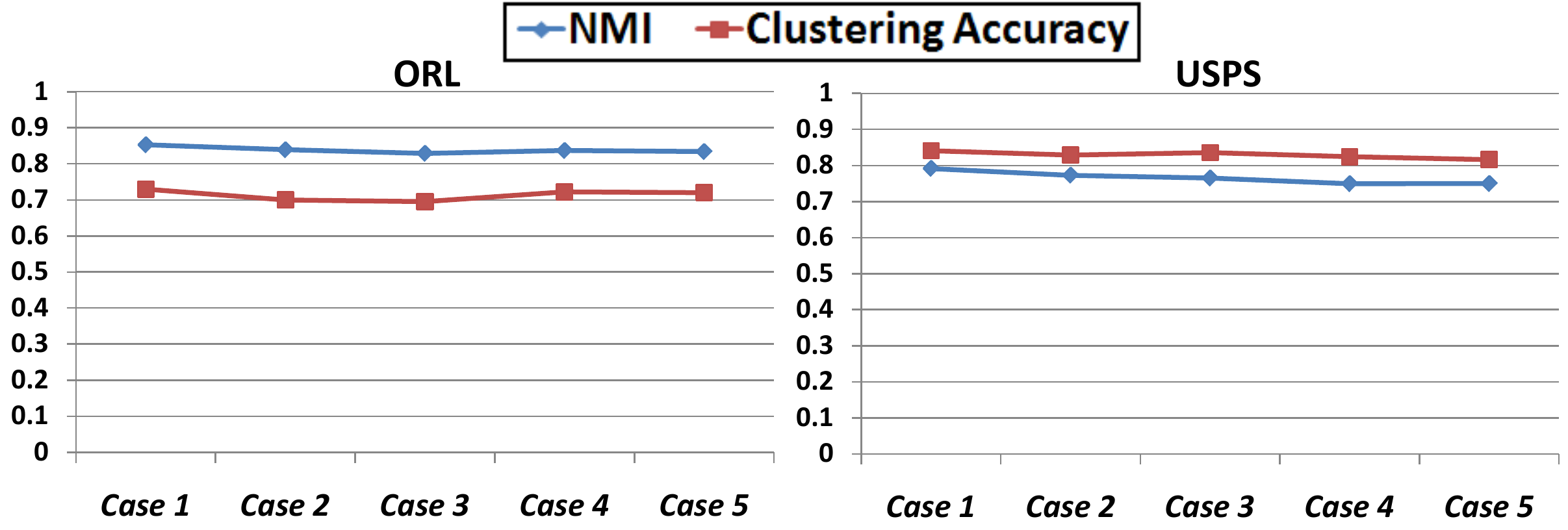}
\end{center}
\vspace{-0.19cm}
\caption{Clustering performances of the proposed PKO+HSC with different
numbers of vertex communities (i.e., $\{2\mathcal{K}, 3\mathcal{K}, 4\mathcal{K}, 5\mathcal{K}, 6\mathcal{K}\}$
with $\mathcal{K}$ being the desired number of clusters)
for high-order over-clustering
hypergraph construction in NMI and accuracy on the two datasets without corruption.
The $(\alpha, \beta)$ configurations for weighting different types of hypergraphs on the two  datasets are
as follows: ORL$\rightarrow$(0.6, 0.2) and USPS$\rightarrow$(0.4,0.4).
}
\label{fig:OverClustering} \vspace{-0.6cm}
\end{figure}

Besides, we report the clustering results of the proposed PKO+HSC with different configurations
of the nearest-neighbor number $k$ for \KNN hypergraph construction
in Fig.~\ref{fig:ORL_K_Selection}.
From Fig.~\ref{fig:ORL_K_Selection}, we see that the proposed PKO+HSC
is not very sensitive to the settings of $k$.
Moreover, Fig.~\ref{fig:OverClustering} shows the clustering performances of
the proposed PKO+HSC using different numbers of vertex communities
for high-order over-clustering hypergraph construction in NMI and clustering
accuracy on the two datasets.
It is clearly seen from Fig.~\ref{fig:OverClustering} that
the proposed PKO+HSC is not very sensitive to the choice of
the vertex community number.

Overall, the proposed PKO+HSC outperforms the other clustering algorithms.
Considering three types of hypergraph information (i.e., pairwise, \KNN, and over-clustering),
the proposed PKO+HSC is capable of effectively exploring the intrinsic topological information among vertices.
By optimizing the discriminative hypergraph partitioning criterion (DHPC),  the proposed PKO+HSC
considers both intra-cluster compactness and inter-cluster separability, resulting in
the overall clustering robustness.

\vspace{-0.1cm}
\section{Conclusion and future work\label{sec:Conclusion}}
\vspace{-0.06cm}

In this work, we have proposed a context-aware hypergraph similarity
measure (CAHSM), which is based on  three types of
hypergraphs---pairwise hypergraph, $k$-nearest-neighbor (\KNN)
hypergraph, and high-order over-clustering hypergraph. These hypergraphs capture the
pairwise, neighborhood, and local grouping information on vertices. By
effectively combining these types of affinity information, CAHSM is capable of effectively exploring the
intrinsic structural information on vertices, resulting in the
robust clustering performance.  In order to fully capture the
intra-cluster compactness and the inter-cluster separability of
vertices, we have also designed a discriminative hypergraph partitioning
criterion (DHPC) that is solved by trace-ratio maximization.
Based on both CAHSM and DHPC, a robust spectral clustering
algorithm (referred to as PKO+HSC) is developed for data
clustering.  Experimental results on various datasets, with and without
noisy perturbation and outlier corruption, demonstrate that the
proposed PKO+HSC has higher clustering robustness and
effectiveness than competing algorithms in most cases.

On the other hand, this work is likely to have two limitations: 1)
it is incapable of adaptively combining the aforementioned three types of hypergraphs;
and 2) the number of clusters in spectral clustering is required to be
provided in advance. Therefore, our future work is to figure out
an adaptive weighting mechanism for hypergraph combination
and an effective scheme for automatically estimating
the cluster number prior to spectral clustering.

\begin{table}
\scriptsize
\scalebox{0.88}{
\begin{tabular}{c|c|c|c|c|c|c|c|c}
\hline
& \makebox[0.9cm]{PKO+HSC} & \makebox[0.5cm]{PK+HSC} & \makebox[0.5cm]{PO+HSC} & \makebox[0.5cm]{PKO+NC} & \makebox[0.5cm]{CSC}
 & \makebox[0.5cm]{NRSC} & \makebox[0.5cm]{RWDSM} & \makebox[0.5cm]{STSC}  \\
\hline
ORL	    & \bf 0.7670	&0.7376	&0.7472	&0.7341	&0.7329	&0.6612	&0.7143	&0.6731\\
YaleB	&\bf 0.4959	&0.4762	&0.4852	&0.4793	&0.4642	&0.3960	&0.4496	&0.4178\\
USPS	&\bf 0.6814	&0.6135	&0.6548	&0.6703	&0.6093	&0.4459	&0.5726	&0.4232\\
MNIST	&\bf 0.5220	&0.4591	&0.5081	&0.5110	&0.4152	&0.3862	&0.4113	&0.3747\\
\hline\hline
ORL	    &\bf 0.6300	&0.6019	&0.6030	&0.5933	&0.5805	&0.5093	&0.5632	&0.5418\\
YaleB	&\bf 0.3701	&0.3311	&0.3597	&0.3612	&0.3346	&0.2575	&0.2946	&0.3010\\
USPS	&\bf 0.7420	&0.6969	&0.7183	&0.7317	&0.6861	&0.5133	&0.6549	&0.4554\\
MNIST	&\bf 0.5985	&0.5497	&0.5740	&0.5883	&0.5244	&0.4585	&0.5398	&0.4529\\
\hline
\end{tabular}
}
\caption{Clustering performances of the eight clustering algorithms by averaging different outlier corruption levels.
Specifically, the upper part corresponds to NMI; and the lower part is associated with accuracy.
The $(\alpha, \beta)$ configurations for weighting different types of hypergraphs on the four datasets are
as follows: (ORL, YaleB.)$\rightarrow$(0.6, 0.2) and (USPS, MNIST)$\rightarrow$(0.4,0.4).
\label{Tab:face_and_digit_noise_v3} \vspace{-0.19cm}}
\end{table}

\end{spacing}

{\footnotesize

\vspace{-0.13cm}
\begin{spacing}{1.0}

\end{spacing}
}


\begin{thebibliography}{1}


\bibitem{Alpert-Kahng1}
C. J. Alpert and A. B. Kahng,
\newblock
``Multiway Partitioning via Geometric Embeddings, Orderings and
Dynamic Programming,''
\newblock
{\it  IEEE Trans. Computer-aided Design of Integrated Circuits
and Systems,}
\newblock
Vol.14, Iss.11, pp.1342-1358, 1995. \par


\bibitem{Chan-Schlag-Zien2}
P. K. Chan, M. D. F. Schlag and J. Y. Zien,
\newblock
``Spectral K-Way Ratio-Cut Partitioning and Clustering,''
\newblock
{\it  IEEE Trans. Computer-aided Design of Integrated Circuits
and Systems,}
\newblock
Vol.13, Iss.9, pp.1088-1096, 1994. \par


\bibitem{Hendrickson-Leland3}
B. Hendrickson and R. Leland,
\newblock
``An Improved Spectral Graph Partitioning Algorithm for Mapping
Parallel Computations,''
\newblock
{\it  SIAM J. Sci. Comput.,}
\newblock
Vol.16, Iss.2, pp.452-459, 1995. \par

\bibitem{Shi-Malik4}
J. Shi and J. Malik,
\newblock
``Normalized Cuts and Image Segmentation,''
\newblock
{\it  IEEE Trans. Pattern Aanalysis  Mach. Intelli.,}
\newblock
Vol.22, Iss.8, pp.888-905, 2000. \par


\bibitem{Malik-Belongie-Leung-Shi5}
J. Malik, S. Belongie, T. Leung, and J. Shi,
\newblock
``Contour and Texture Analysis for Image Segmentation,''
\newblock
{\it  Int. J. Computer Vision,}
\newblock
2001. \par







\bibitem{Weiss-ICCV1999}
Y. Weiss,
\newblock
``Segmentation Using Eigenvectors: A Unifying View,''
\newblock
in {\it Proc. Int. Conf. Computer Vision,}
\newblock
pp.975-982, 1999. \par



\bibitem{Meila-Shi-NIPS2000}
M. Meila and J. Shi,
\newblock
``Learning Segmentation by Random Walks,''
\newblock
{\it Proc. Advances in Neural Information Processing Systems,}
\newblock
pp.873-879, 2000. \par



\bibitem{Gdalyahu-Weinshall-Werman7}
Y. Gdalyahu, D. Weinshall and M. Werman,
\newblock
``Self-Organization in Vision: Stochastic Clustering for Image
Segmentation, Perceptual Grouping, and Image Database
Organization,''
\newblock
{\it  IEEE Trans. Pattern Analysis Mach. Intelli.,}
\newblock
Vol. 23,  Iss. 10, pp.1053-1074, Oct. 2001. \par


%

\bibitem{Ding-He9}
C.H.Q. Ding, X. He, H. Zha, M. Gu and H.D. Simon,
\newblock
``A Min-Max Cut Algorithm for Graph Partitioning and Data
Clustering,''
\newblock
in {\it Proc. Int. Conf. Data Mining}, pp.107-114, 2001. \par




\bibitem{Ng-Jordan-Weiss15}
A. Y. Ng, M. I. Jordan and Y. Weiss,
\newblock
``On Spectral Clustering: Analysis and An Algorithm,''
\newblock
{\it  Proc. Advances in Neural Information Processing Systems,}
\newblock
MIT Press, 2001. \par


\bibitem{Zelnik-Manor-Perona-NIPS2005}
L. Zelnik-Manor and P. Perona,
\newblock
``Self-Tuning Spectral Clustering,''
\newblock
{\it Proc. Advances in Neural Information Processing Systems,}
\newblock
pp.1601-1608, 2005. \par








\bibitem{Yu-Shi14}
S.X. Yu and J. Shi,
\newblock
``Multiclass Spectral Clustering,''
\newblock
in {\it Proc. Int. Conf. Computer Vision,}
\newblock
Vol.1, pp.313-319, 2003. \par


%
%
%
%
%
%


\bibitem{Nadler-Lafon-Coifman-Kevrekidis-NIPS2006}
B. Nadler, S. Lafon, R. Coifman, and I. Kevrekidis,
\newblock
``Diffusion Maps, Spectral Clustering and Eigenfunctions of
Fokkerplanck Operators,''
\newblock
{\it Proc. Advances in Neural Information Processing Systems,}
\newblock
pp.955-962, 2006. \par


\bibitem{Chang-Yeung-ICCV2005}
H. Chang and D. Y. Yeung,
``Robust Path-Based Spectral Clustering with Application to Image Segmentation,''
in {\it Proc. Int. Conf. Computer Vision,} 2005.


\bibitem{Li-Liu-Chen-Tang-ICCV2007}
Z. Li, J. Liu, S. Chen, and X. Tang,
\newblock
``Noise Robust Spectral Clustering,''
\newblock
in {\it Proc. Int. Conf. Computer Vision,}
\newblock
2007. \par


%
%
%
%
%
%
%


%
%
%
%


%
%
%
%


%
%
%
%


\bibitem{Von-Luxburg-SC2007}
U. Von Luxburg,
``A Tutorial on Spectral Clustering,'' {\it Statistics and Computing},
Vol. 17,  Iss. 4,  pp. 395-416, 2007.


%
%
%
%
%


\bibitem{Zhou-Huang-Schokopf-nips2006}
D. Zhou, J. Huang, and B. Sch\"{o}kopf, ``Learning With Hypergraphs:
Clustering, Classification, and Embedding,'' in {\it Proc.
Advances in Neural Information Processing Systems}, 2006.


%
%
%

\bibitem{Agarwal-Lim-Manor-Perona-Kriegman-Belongie-CVPR2005}
S. Agarwal, J. Lim, L. Zelnik Manor, P. Perona, D. Kriegman,
and S. Belongie, ``Beyond Pairwise Clustering,'' in {\it Proc.
IEEE Conf. Computer Vision Pattern Recognition}, 2005.

\bibitem{Zass-Shashua-CVPR2008}
R. Zass and A. Shashua, ``Probabilistic Graph and Hypergraph
Matching,'' in {\it Proc. IEEE Conf. Computer Vision Pattern Recognition}, 2008.


\bibitem{Sun-Ji-Ye-KDD2008}
L. Sun, S. Ji, and J. Ye, ``Hypergraph Spectral Learning for
Multi-Label Classification,'' in {\it Proc. ACM SIG KDD}, 2008.


\bibitem{Huang-Liu-Metaxas-CVPR2009}
Y. Huang, Q. Liu, and D. Metaxas, ``Video Object Segmentation
by Hypergraph Cut,'' in {\it Proc. IEEE Conf. Computer Vision Pattern Recognition}, 2009.


\bibitem{Huang-Liu-Zhang-Metaxas-CVPR2010}
Y. Huang, Q. Liu, S. Zhang, and D. N. Metaxas,
``Image Retrieval via Probabilistic Hypergraph Ranking,''
in {\it Proc. IEEE Conf. Computer Vision Pattern Recognition}, 2010.


%
%
%

\bibitem{Oliva-Torralba-IJCV2001-gist}
A. Oliva and A. Torralba, ``Modeling The Shape of The Scene:
A Holistic Representation of The Spatial Envelope,'' {\it Int. J. Computer Vision}, 42(3):145-175, 2001.

\bibitem{Naftel-Khalid-ICVS2006}
A. Naftel and S. Khalid, ``Motion Trajectory Learning in the DFT-Coefficient Feature Space,'' in {\it Proc.
IEEE Int. Conf. Computer Vision Systems}, 2006.


\bibitem{Lauer-Schnorr-iccv2009}
F. Lauer and C. Schn\"{o}rr,
``Spectral Clustering of Linear Subspaces for Motion Segmentation,'',
in {\it Proc. Int. Conf. Computer Vision}, 2009.

\bibitem{Hu-Xie-TIP2007}
W. Hu, D. Xie, Z. Fu, W. Zeng, and S. Maybank,
``Semantic-Based Surveillance Video Retrieval,''
{\it IEEE Trans. on Image Processing}, Vol. 16, Iss. 4, pp. 1168-1181, 2007.

\bibitem{Hsieh-Yu-Chen-TCSVT2006}
J. Hsieh, S. Yu, and Y. Chen,
``Motion-Based Video Retrieval by Trajectory Matching,''
{\it IEEE Trans. on Circuit System for Video Technology}, Vol. 16, Iss. 3, pp. 396-409, 2006.

\bibitem{Wang-Shen-Zheng-Ren-ACCV2009}
P. Wang, C. Shen, H. Zheng, and Z. Ren,
``A Variant of the Trace Quotient Formulation for
Dimensionality Reduction,'' pp. 277-286,
in {\it Proc. Asian Conf. Computer Vision}, 2009.

\bibitem{Wang-Yan-Xu-Tang-Huang-CVPR2007}
H. Wang, S. Yan, D. Xu, X. Tang, and T. Huang,
Trace Ratio vs. Ratio Trace for Dimensionality Reduction,
in {\it Proc. IEEE Conf. Computer Vision Pattern Recognition}, 2007.


\bibitem{Jia-Nie-Zhang-TNN2009}
Y. Jia, F. Nie, and C. Zhang,
Trace Ratio Problem Revisited,
{\it IEEE. Trans. on Neural Networks}, Vol. 20, Iss. 4
pp. 729-735, 2009.

%
%
%
%

\bibitem{lu2008constrained}
Z. Lu and M.A. Carreira-Perpi\~{n}\'{a}n,
``Constrained Spectral Clustering through Affinity Propagation,''
in {\it Proc.  IEEE Conf. Computer Vision Pattern Recognition}, 2008.


\bibitem{Li-ACCV2009-SC}
X. Li, W. Hu, Z. Zhang, and Y. Liu,
``Spectral Graph Partitioning Based on a Random Walk
Diffusion Similarity Measure,''
in {\it Proc. Asian Conf. Computer Vision}, 2009.

%
%
%
%

\bibitem{Ngo-SIAM2012-traceratio}
T. T. Ngo, M. Bellalij, and Y. Saad,
``The Trace Ratio Optimization Problem,''
{\it SIAM Review}, Vol. 54, Iss. 3,
pp. 545-569, 2012.


\bibitem{yang2012discriminative}
Y. Yang, Y. Yang, Y. Zhang, X. Du, H. T. Shen, and X. Zhou,
``Discriminative Nonnegative Spectral Clustering with Out-of-Sample Extension,''
{\it IEEE Trans. Knowledge and Data Engineering},
2012.


\bibitem{wang2009clustering}
F. Wang, C. Zhang, T. Li,
``Clustering with Local and Global Regularization,''
{\it IEEE Trans. Knowledge and Data Engineering},
 Vol. 21, Iss. 12, pp. 1665-1678, 2009.


\bibitem{cai2007spectral}
D. Cai, X. He, and J. Han,
``Spectral Regression for Efficient Regularized Subspace Learning,''
in {\it Proc. Int. Conf. Computer Vision}, 2007.



\bibitem{Lanczos_wiki}
http://en.wikipedia.org/wiki/Lanczos\_algorithm


\bibitem{shen2008supervised}
C. Shen, H. Li, and B. J. Michael,
``Supervised Dimensionality Reduction via Sequential Semidefinite Programming,''
{\it Pattern Recognition}, Vol. 41, Iss. 12, pp. 3644-3652, 2008.

\end{thebibliography}
\end{document}